\documentclass[sigconf, 10pt]{acmart} 

\usepackage{subcaption}
\usepackage{booktabs} 
\usepackage[linesnumbered,vlined,ruled]{algorithm2e}
\usepackage{amsmath,amssymb,amsfonts}

\usepackage{enumitem}
\setcopyright{rightsretained}
\usepackage{mathtools}
\usepackage{multirow}
\usepackage{tikz}

\newif\ifstartedinmathmode
\newcommand\encircled[1]{%
  \relax\ifmmode\startedinmathmodetrue\else\startedinmathmodefalse\fi%
  \tikz[baseline,anchor=base]{%
  \node[draw,circle,outer sep=0pt,inner sep=.2ex]
    {\ifstartedinmathmode$#1$\else#1\fi};}%
}

\acmConference[]{MobiXXX 2020}{September 2020}{London, UK}
\acmYear{2020}
\copyrightyear{2020}

\begin{document}
	

\title{Conservative Plane Releasing for Spatial Privacy Protection in Mixed Reality}

\author{Jaybie Agullo de Guzman}
\orcid{0002-2816-7721}
\affiliation{%
  \institution{Data 61|CSIRO \&\\
  University of New South Wales}
  \city{Sydney}
\country{Australia}
}
\email{jaybie.deguzman@data61.csiro.au}

\author{Kanchana Thilakarathna}
\affiliation{%
  \institution{University of Sydney}
  \city{Sydney}
	\country{Australia}
}
\email{kanchana.thilakarathna@sydney.edu.au}

\author{Aruna Seneviratne}
\affiliation{%
  \institution{University of New South Wales}
  \city{Sydney}
  \country{Australia}
}
\email{a.seneviratne@unsw.edu.au}

\renewcommand{\shortauthors}{J. de Guzman, K. Thilakarathna, \& A. Seneviratne}

\begin{abstract}
Augmented reality (AR) or mixed reality (MR) platforms require spatial understanding to detect objects or surfaces, often including their structural (i.e. spatial geometry) and photometric (e.g. color, and texture) attributes, to allow applications to place virtual or synthetic objects seemingly ``anchored'' on to real world objects; in some cases, even allowing interactions between the physical and virtual objects. These functionalities require AR/MR platforms to capture the 3D spatial information with high resolution and frequency; however, these pose unprecedented risks to user privacy. Aside from objects being detected, spatial information also reveals the location of the user with high specificity, e.g. in which part of the house the user is. In this work, we propose to leverage \textit{spatial generalizations} coupled with \textit{conservative releasing} to provide spatial privacy while maintaining data utility. We designed an adversary that builds up on existing place and shape recognition methods over 3D data as attackers to which the proposed spatial privacy approach can be evaluated against. Then, we simulate user movement within spaces which reveals more of their space as they move around utilizing 3D point clouds collected from Microsoft HoloLens. Results show that revealing no more than 11 \textit{generalized} planes--accumulated from successively revealed spaces with large enough radius, i.e. $r\leq1.0m$--can make an adversary fail in identifying the spatial location of the user for at least half of the time. Furthermore, if the accumulated spaces are of smaller radius, i.e. each successively revealed space is $r\leq 0.5m$, we can release up to 29 \textit{generalized} planes while enjoying both better data utility and privacy.
\end{abstract}

%
%




\maketitle

\section{Introduction}

\begin{figure}[t]
	\centering
	\includegraphics[width=\columnwidth]{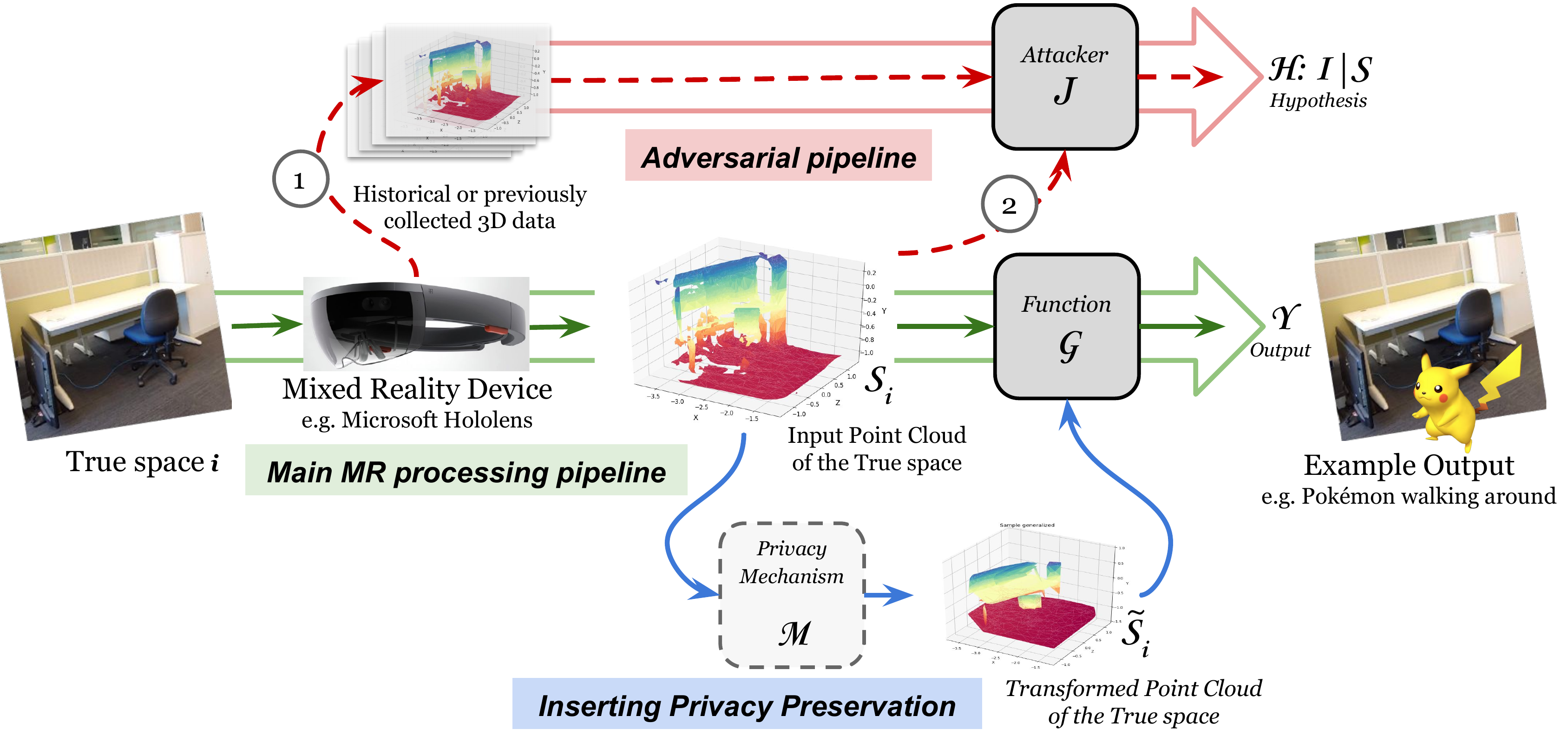}
	\vspace{-5mm}
    \caption{\small Information flow diagram (following the green solid arrows) for an intended MR function $G$, with an attacker $J$ which can perform adversarial spatial inference: (1) adversarial inference \textit{modeling} or \textit{learning} from, say, historical 3D data, and (2) adversarial inference or \textit{matching} over currently released 3D data. Then, inserting an intermediate privacy-preserving mechanism $M$ which aims to prevent adversarial spatial inference.}
	\label{fig:adversary-model-pipeline}
	\vspace{-5mm}
\end{figure}


AR/MR platforms such as Google ARCore, 
Apple ARKit, 
and Windows Mixed Reality API 
requires spatial understanding of the user environment in order to deliver virtual augmentations that seemingly inhabit the real world, and, in some immersive examples, even interact with physical objects.\footnote{ARKit, \url{https://developer.apple.com/documentation/arkit};
ARCore, \url{https://developers.google.com/ar/};
Windows MR, \url{https://www.microsoft.com/en-au/windows/windows-mixed-reality}. For the rest of the paper, we will be collectively referring to AR and MR as MR.} Fig. \ref{fig:adversary-model-pipeline} shows a generic information flow diagram for MR. The captured spatial information is stored digitally as a \textit{spatial} map or graph of 3D points, called a \textit{point cloud} (labelled $S_i$ in Fig. \ref{fig:adversary-model-pipeline}), which is accompanied by mesh information to indicate how the points, when connected, represent surfaces and other structures in the user environment. 
However, these 3D spatial maps that may contain sensitive information, which the user did not intend to expose, can 
be further utilized for functionalities beyond the application's intended function (see potential attacker $J$ in Fig. \ref{fig:adversary-model-pipeline}) for benign attacks such as aggressive localized advertisements to malevolent ones such as burglaries. Nonetheless, there are no mechanisms in place that ensure user \textit{spatial} data privacy in existing MR platforms.

Moreover, despite 3D data being a structural representation of the real world, 3D data is perceptually latent from the users. With traditional media, such as images and video, what the ``machine sees'' is what the ``user sees''. On the other hand, with MR, what the machine sees is different--often even more--than what the user sees. With MR, the experience is exported as visual data (e.g. objects augmented on the user's view) while the user is oblivious about the captured spatial mapping, its resolution, and exactness. This inherent perceptual difference creates a latency from user perception and, perhaps, affects--or the lack thereof--how users perceive the sensitivity of 3D information. Aside from the spatial structural information, the mapping can also include 3D maps of objects within the space. Furthermore, the spatial map can also reveal the location of the user: both the \textit{general} location, and their specific location \textit{within} the space. Moreover, most users are oblivious about the various information that are included in the spatial maps captured and stored by MR platforms.

As majority of the work on MR have been focused on delivering the technology, 
only recently have there been efforts in addressing the security, safety, and privacy risks associated with MR technology \cite{deguzman2018security, deguzman2019firstlook}. There have been a few older works that have pointed out the issues on ethical considerations \cite{heimo2014ethical} in MR as well as highlighting considerations on data ownership, privacy, secrecy, and integrity \cite{friedman2000value}. Moreover, potential perceptual and sensory threats that can arise from MR outputs such as photosensitive epilepsy and motion-induced blindness have also been looked into \cite{baldassi2018challenges}. In conjunction to these expositions, the EU have recently legislated the General Data Protection Regulation (GDPR) which aims to empower users and protect their data privacy through a policy approach. This further highlights the importance of designing and developing \textit{privacy-enhancing technologies} (PETs) especially those that can be applied to the MR use case.

In light of this, first, we present two adversary approaches that  
recognizes the general space, i.e. \textit{inter}-space, and also infers the user's location within the space, i.e. \textit{intra}-space. To construct these attackers, we build up on existing 3D place recognizers that have been applied over 3D lidar data: an intrinsic descriptor matching-based 3D object recognizer, and a deep neural network-based 3D shape classifier. We modify them to operate in the scale on which 3D data is captured by MR platforms. We demonstrate how easy it is to extend these methods to be used as an attacker in the MR scenario whilst quantifying the privacy leakage. 
Then, we propose \textit{spatial plane generalizations} with \textit{conservative plane releasing} as a privacy-enhancing approach which can potentially be easily integrated with existing MR platforms--i.e. inserted as an intermediary privacy mechanism as shown in Fig. \ref{fig:adversary-model-pipeline}. Finally, we evaluate not only the privacy leakage, but also the reduction of utility in terms of quality of service utilizing real-world 3D point cloud data captured through the Microsoft HoloLens from a variety of spaces. To this end, we summarize the major contributions as follows:

\begin{enumerate}[leftmargin=*]
	\item We present a \textit{3D adversarial inference model} that reveals the general space of a user, i.e. \textit{inter}-space inference, and their specific location within the space, i.e. \textit{intra}-space inference.
	\item We compare the performance of the two spatial inference attack approaches and show that a `classical' descriptor-based matcher can outperform a deep neural network-based recognizer.
	\item We demonstrate that the insufficient protection provided by spatial \textit{generalizations} can be improved by \textit{conservatively} releasing the plane generalizations; specifically, controlling the maximum number of released generalized planes instead of naively providing the generalizations entirely. 
	\item We present an in depth analysis over a realistic scenario when user spaces are successively released and experimentally determine the maximum number of releases, and generalized planes that prevents spatial inference, both inter-space and intra-space. For example, revealing no more than 11 generalized planes can make an adversary fail in identifying the spatial location of the user for at least half of the time.
	\item Lastly, we show that better data utility can be achieved with a smaller size, i.e. r = 0.5m, of revealed generalized spaces while providing adequate--or even better--privacy; specifically, we can release up to 29 generalized planes while enjoying both better data utility and privacy.
\end{enumerate}

The rest of the paper is organized as follows. First, we discuss the related work in \S\ref{sec:related-work}, and present the theoretical framework of the spatial privacy problem, and associated definitions in \S\ref{sec:framework}. Then, in \S\ref{sec:inference}, we describe two attack methods an adversary may utilize for spatial inference. In \S\ref{sec:privacy-preservation}, we describe the various information reduction techniques we can employ to prevent spatial inference. We present the evaluation methodology in \S\ref{sec:methodology} to investigate the viability of privacy methods we employ over varying experimental setups. The results are presented in \S\ref{sec:results} followed by the discussion in \S\ref{sec:discussion}. Lastly, we conclude the paper in \S\ref{sec:conclusion}. 

\section{Related Work}\label{sec:related-work}

In the general privacy and security space, various work have already been done: from revealing privacy leakage in online social networks \cite{krishnamurthy2010privacy} and mobile devices \cite{ren2016recon}, to developing privacy-preserving mechanisms for conventional user data types \cite{sweeney2002k, mcsherry2007mechanism, dwork2014algorithmic} as well as against deep-learning inference \cite{shokri2015privacy, abadi2016deep}. 
\vspace{-3mm}
\paragraph{\textbf{Information Sanitization.}} Consequently, several AR/MR and related PETs have been proposed in the literature, and are listed in \cite{deguzman2018security}. Early approaches primarily involved applying sanitization techniques on visual media (i.e. image and video): e.g. removing RGB and only showing contours \cite{jana2013scanner}, detecting markers that signify sensitive content to be sanitized \cite{raval2014markit, raval2016you},  sanitizing sensitive objects from a shared database or communicated through PAN \cite{roesner2014world, aditya2016pic, li2016privacycamera}, through user gestures \cite{shu2016cardea}, or based on context \cite{zarepour2016context, steilPrivaceye2018}. However, these sanitization approaches are focused on post-captured images which can still pose security and privacy risks, and they have only been applied on the wide use case of visual capture devices and not on actual MR devices or platforms.

\vspace{-4mm}
\paragraph{\textbf{Visual information abstraction.}} Abstraction addresses the earlier issues by reducing or eliminating the necessity of accessing the raw visual feed directly. In the specific 3D use case, significant work have been done on protections involving abstracted physiological information \cite{jana2013enabling, figueiredo2016prepose} using the idea of \textit{least privilege} \cite{vilk2014least}. The same approach has also been used for providing spatial information while maintaining visual privacy \cite{vilk2015surroundweb}. A recent work also utilized the idea of least privilege to apply visual abstractions for mobile MR \cite{deguzman2019safeMR}. However, they are functionality or data specific, and have neither presented or exposed actual risks with 3D MR data nor provided evaluation against attacks.

\vspace{-4mm}
\paragraph{\textbf{3D data: attacks, risks, and protection.}} Very few recent works have started to look into the actual risks brought about by indefinite access to 3D data. One provided preliminary evidence of how an adversary can easily infer spaces from captured 3D point cloud data from Microsoft Hololens \cite{deguzman2019firstlook}; and how, even with spatial generalization (i.e. the 3D space is generalized in to a set of planes), spatial inference is still possible at a significant success rate; however, they did not use a machine learning approach which may further demonstrate the ineffectiveness of spatial generalizations as privacy protection. In contrast, another recent work employed machine learning to reveal original scenes from 3D point cloud data with additional visual information \cite{pittaluga2019revealing}. While a concurrent work focused on a privacy-preserving method of pose estimation to counter the scene revelation \cite{speciale2019privacy}: 3D ``line'' clouds are used during pose estimation to obfuscate 3D structural, i.e. geometrical, information; however, this approach only addresses the pose estimation functionality and does not present the viability for surface or object detection which is necessary for a virtual object to be rendered or ``anchored'' onto. Thus, it is still necessary for 3D point cloud data to be exposed but with privacy-preserving transformations to hide sensitive content and prevent spatial recognition. 

\vspace{2mm}
In this work, we improve upon the attacker used in \cite{deguzman2019firstlook} by extending the attack with intra-space inference, and adding a deep neural network-based approach in our investigation as a potentially stronger attacker. Moreover, we augment the currently inadequate spatial generalizations with conservative releasing as a stronger countermeasure against these attacks.

\section{Spatial Privacy Problem in Mixed Reality}\label{sec:framework}


As shown in Fig. \ref{fig:adversary-model-pipeline} and following the notation map in Tab. \ref{tab:notation-map}, we define a space represented by a point cloud $S$ identified by a label $i$ which can be segmented into a set of overlapping point cloud subspaces $S = \{s_1, s_2, ... s_n\}$. An MR functionality $G$ produces an output $Y$, and from which we derive the utility or QoS function $Q$. An adversarial inferrer $J$ produces a hypothesis $H$ to reveal the spatial location of an unknown space. Lastly, a privacy preserving mechanism $M$ transforms $S$, or its subspaces $s \subset S$, to a privacy-preserving version $\widetilde{S} \text{ or } \tilde{s}$, i.e. $M : S \text{ or } s\mapsto \widetilde{S} \text{ or }\tilde{s}$. 

\renewcommand{\arraystretch}{1.15}
\begin{table}[t]
	\caption{Notation Map}
	\label{tab:notation-map}
    \vspace{-3mm}
	\centering
	\resizebox{0.93\columnwidth}{!}{
	\begin{tabular}{@{}cl@{}}
		\toprule
		\textbf{Notation} & \textbf{Description} \\ \midrule
		$S_i$ & point cloud representation of space labelled $i$; \\
		& \hspace{5mm} can be composed of subspaces  $s$, $ s \subset S$; and  \\
		& \hspace{5mm} composed of oriented points $p$, $\forall p \in s \mapsto \forall p \in S$   \\
		$M$ & privacy-preserving mechanism             \\
		$\widetilde{S}_{(i),v}$ & transformed point cloud released by $M$            \\
		$G$ & intended functionality (i.e. MR app or service)             \\
		$Y_{i,v}$ & output of the intended functionality $G$     \\ 
		$Q_{\widetilde{S};S}$ & difference of the transformed $\widetilde{S}$ from $S$  \\ 	
		$J$ & adversarial inferrer \\
		$h$ & hypothesis of $J$ about an unknown query space $i^*$;  \\	
		& \hspace{5mm} producing an inter-space hypothesis $i^* = i$, and\\
		& \hspace{5mm} an intra-space hypothesis centroid $c_S$ \\
$\Pi_1(\widetilde{S};S)$ & inter-space privacy measure in terms of classification error  \\
		$\Pi_2(\widetilde{S};S)$ & intra-space privacy measure in terms of distance error  \\
		\bottomrule
	\end{tabular}}
	\vspace{-7mm}
\end{table}

In the succeeding sections, we formalize the adversary model in \S\ref{subsec:adversary-model}, the privacy metrics in \S\ref{subsec:privacy}, and the functionality and utility metrics in \S\ref{subsec:mr-functionality}. Before we proceed, we first introduce 3D data representation.

\subsection{3D spatial data}\label{subsec:3d-data}


There are various ways that MR-capable devices capture and compute 3D spatial data. Depending on the platform, the underlying environmental mapping approach would be any or combinations of the following: simultaneous localization and mapping (SLAM), visual odometry, and/or structure from motion (SfM). We direct the readers to \cite{saputra2018visual} for a survey on these visual mapping algorithms.

Regardless of the underlying mapping algorithm, the aim is to construct a 3D spatial map represented by a set of \textit{oriented points}: 3D points described by their $\{x, y, z\}$-position in space and usually accompanied by a normal vector $\{n_x, n_y, n_z\}$ which informs of the orientation of the surface to which the point belongs to. (When normal vectors are not readily available, it is estimated from the points themselves.) 
A collection of these oriented points together with an accompanying mesh information, which informs of how the points are connected to form surfaces, constitute a \textit{point cloud}.


\subsection{Adversary model}\label{subsec:adversary-model}
Currently, there are no mechanisms in place that ensures user data privacy on MR platforms.
As shown in Fig. \ref{fig:adversary-model-pipeline}, any potentially adversarial application can access and store all captured 3D spatial maps. These adversaries may desire to infer the location of the users, or they may further infer user poses, movement in space, or even detect changes in user environment. Furthermore, in contrast to video and image capture, 3D data can provide a much more lightweight and near-truth representation of user spaces.

In this work, we will focus on the spatial inference attack where the adversary aims to recognize the location of the user on \textit{two} levels given historical 3D data of user spaces: 
(1) the \textit{general} location of the user (among the known ensemble of spaces) we call \textit{inter-space} inference, and (2)
the \textit{specific} location of the user within the space we call \textit{intra-space} inference. 
Also, we assume that the adversary's aim is to individually infer user spaces; thus, an attacker will develop a different reference model for every user. And that it can only infer spaces that the user has historically been in.


\paragraph*{Defining adversarial inference} As shown in Fig. \ref{fig:adversary-model-pipeline}, inference is a two-step process: (1) the training of a \textit{reference} model or creation of a dictionary using 3D description algorithms over the previously known spaces as reference, (2) and the inference of unknown spaces by testing the model, i.e. \textit{matching} their 3D descriptors to that of the reference descriptors from step 1. We assume that the adversary has \textit{prior knowledge} about the spaces which they can use as reference. Prior knowledge can be made available through \textit{historical} or publicly available 3D spatial data of the user spaces, \textit{previously provided} data by the user themselves or other users, or from a \textit{colluding application} or service that has access to raw or higher resolution 3D data. Furthermore, we assume that the adversary is aware of the generalizations that can be applied on released point cloud data and be able to adjust its attack accordingly.

Then, an attacker $J$ produces a two-element hypothesis $h$ for our two-level attack:  \textit{inter-space} location, i.e.  location $i^*=i$, to determine in which of the reference spaces the query space is; and \textit{intra-space} location defined by a centroid $c_{S^*} = \{x^*,y^*,z^*\}$ of an unknown point cloud ${S_i^*}$.
Hypothesis $h$ is defined as follows:
\begin{align}
\footnotesize J:S_i^* \longrightarrow h &: i^*=i \text{, and } c_{s^*} = \{x,y,z\} \text{, where} \label{eq:inference}\\
\footnotesize i^*=i &= \underset{\forall i}{\text{ argmax }}\ L_{S^*}(i^*=i) \text{, and} \tag{a} \\
\footnotesize c_{S^*} &= \{x^*,y^*,z^*\} \mapsto c_{S_r} = \{x_r,y_r,z_r\}, \tag{b} 
\end{align}
where $L_{S^*}(i^* = i)$ is a likelihood function about the unknown space $S^*$ having a label $i^*=i$, and $\{x_r,y_r,z_r\} = |\{k_{S_r,b}\}|_{centroid}$ is the centroid of the best reference key point matches.

Intra-space inference is the primary step in tracking or localization. Aside from knowing the general location of a user, which can have a huge width such as $r\approx100 m$, an attacker would also  require the exact location of the user. For example, if an adversarial advertising service with various points of presence (PoP) can tell if one of their PoP is within user view, then, the service can ship an advertisement to the PoP nearest to the user. Conversely, if the attacker wrongly infers the intra-space location of the user, then the attack has failed. The methods employed for inference are further discussed in \S\ref{sec:inference}. 



\subsection{Privacy Metrics}\label{subsec:privacy}

\paragraph*{Defining the privacy metrics} We define the \textit{inter-space privacy} function $\Pi_1$ in terms of the \textit{inter-space misclassification error rate} of the inferrer $J$. A few works in the literature uses a similar classification error-based metric for privacy \cite{Wagner:2018:TPM:3212709.3168389, shokri2011quantifying}. We treat the correct classification event as a discrete random variable defined by a discrete delta function
$\footnotesize
\delta(i^*,i) = \{ 1 \text{ if } i^* = i; 0 \text{ if } i^* \neq i \}$ so that we can pose $\Pi_1$ as the expectation of misclassification given an unknown point cloud $S^*$ as follows
\begin{equation} 
    \footnotesize \Pi_1(S^*;S) = \sum_{\forall i} \sum_{\forall i^*} (1 - \delta(i^*,i))L_{S^*}(i^* = i) 
	\label{eq:inter-space-privacy}
	\vspace{-1mm}
\end{equation}
where $i^*$ is the hypothesis label of $J$ about the true label $i$ of an unknown space $S^*$, and $L_{s^*}(i^* = i)$ is the likelihood function.

Likewise, we can also pose a secondary metric $\Pi_2$ for \textit{intra-space privacy} which captures how accurate an adversary can estimate a user's position within a space; that is, the user, at any given time is at a specific subspace within the larger known space. We define the \textit{mean intra-space distance error} as follows:
\begin{equation} 
    \footnotesize \Pi_2(\widetilde{S};S) = \frac{1}{|\forall \{S^*,S: i^* = i\}|} \sum_{\forall \{S^*,S: i^* = i\}} d(c_{S^*},c_{S}) 
	\label{eq:intra-space-privacy}
	\vspace{-1mm}
\end{equation}
where the $d(c_{S^*},c_{S})$ represent the total distance error between the correct intra-space location $c_{S} = \{x,y,z\}$ and the hypothesis intra-space location $c_{S^*} = \{x^*,y^*,z^*\} = \{x_r,y_r,z_r\}$ for a given unknown space $S^*$. The distance formula is as follows:
\begin{flalign}
\footnotesize d(c_{S^*},c_{S}) &= ||c_{S^*}-c_S|| \nonumber\\
\footnotesize &= \sqrt{(x_r-x)^2 + (y_r-y)^2+ (z_r-z)^2}. \label{eq:intra-space-distance}
\end{flalign}

\subsection{Mixed reality functionality}\label{subsec:mr-functionality}
Perhaps, the most common functionality in MR is the \textit{anchoring} of virtual 3D objects on to real world surfaces (e.g. the floor, walls, or tables). At the minimum level, a \textit{static} anchor only requires an $\{x,y,z\}$ point with an orientation $\{n_x,n_y,n_z\}$. A set of these oriented points which forms a plane or surface can allow for \textit{dynamic} virtual augmentations that move about the surface. Furthermore, a set of these planes or surfaces can constitute a space which, then, allows for augmentations that move about the space.




\paragraph*{Defining the function utility} For a given functionality $G$, an effective privacy mechanism $M$ aims to make the resulting outputs $Y$ from the raw point cloud $S$ and its privacy-preserving version $\widetilde{S}$ similar, i.e. $Y_S \simeq Y_{\widetilde{S}}$. Thus, we define utility as the \textit{Quality-of-Service} (QoS) in terms of the \textit{output transformation error}, $Q_{Y_{\widetilde{S}};Y_S}$, or the difference of the transformed output  $Y_{\widetilde{S}}$ from the original output  $Y_{S}$: 
$\footnotesize Q_{Y_{\widetilde{S}};Y_S} = |Y_S - Y_{\widetilde{S}}|$,
which we aim to minimize.

For consistently anchored augmentations, some functionalities require near-truth point cloud representations. This implies that we can directly compute the $Q_{Y_{\widetilde{S}};Y_S}$ as the difference $Q_{\widetilde{S};S} $ of the point clouds themselves instead of the outputs like so 
$\footnotesize Q_{\widetilde{S};S}  = |S - \widetilde{S} |.$

\paragraph*{Defining the utility metric} 
We can define $Q_{\widetilde{S};S}$ as a utility metric by specifying it as a \textit{mean transformation error} (QoS) with the following components
\begin{equation}
\footnotesize
Q_{\widetilde{S};S} = \underset{\forall \tilde{p} \in \widetilde{S},\forall p \in S}{\text{mean}} [\alpha \cdot (||p - \tilde{p}||) + \beta \cdot (1-\vec{n_p} \cdot \vec{n}_{\tilde{p}}) ]
\vspace{-1mm}
\label{eq:utility-by-qos}
\end{equation}
where the first component, i.e. $||p - \tilde{p}||$, is the \textit{point-wise} Euclidean difference of the true/raw point $p$ from the transformed point $\tilde{p}$, while the second component are their normal vector difference using the difference from 1 of their normal vector cosine similarity (i.e. $\vec{n_p} \cdot \vec{n}_{\tilde{p}}$). The coefficients $\alpha$ and $\beta$ are contribution weights where $\alpha, \beta \in \left[0,1\right]$ and $\alpha + \beta  = 1$. We set $\alpha, \beta = 0.5$. To compute this point-wise differences, we have to first find the nearest neighbor pairs of each point in the transformed point cloud $\widetilde{S}$ from the raw point cloud $S$, i.e. $\tilde{p} \in \widetilde{S} \overset{1\mathtt{nn}}{\longmapsto} p \in S$; thus, the differences are computed between the $1\mathtt{nn}$ $\{\tilde{p},p\}$-pair of points and is computed $\forall \tilde{p} \in \widetilde{S}$ to get the mean difference over the entire transformed point cloud $\widetilde{S}$.

Moreover, we also specify an inequality constraint $\gamma$ that defines the maximum permissible transformation error as follows:
\begin{equation}
\footnotesize \underset{\forall \tilde{p} \in \widetilde{S},\forall p \in S}{mean} [\alpha \cdot (1 - || \tilde{p} - p||) + \beta \cdot (\vec{n}_{\tilde{p}} \cdot \vec{n}_p) ] \leq \gamma
\label{eq:constrained-qos}\end{equation}
The $\gamma$ can be used as a tunable parameter depending on the exactness required by the MR function: higher gamma implies that the MR function does not require released spaces to be very exact to the true space, while a small gamma implies exactness.

A desired mechanism $M$ produces $\widetilde{S}$ that maximizes the privacy functions $\Pi_1$ and $\Pi_2$ while minimizing $Q$. Moreover, we will freely use the following notation $\Pi_{1}(\Theta)$ or $\Pi_{2}(\Theta)$ where we indicate a set of parameters $\Theta$ that specifies the transformation $M$.

\section{Adversarial Spatial Inference}\label{sec:inference}

\begin{figure}[t]
	\vspace{-2mm}
	\centering
	\includegraphics[width=\columnwidth]{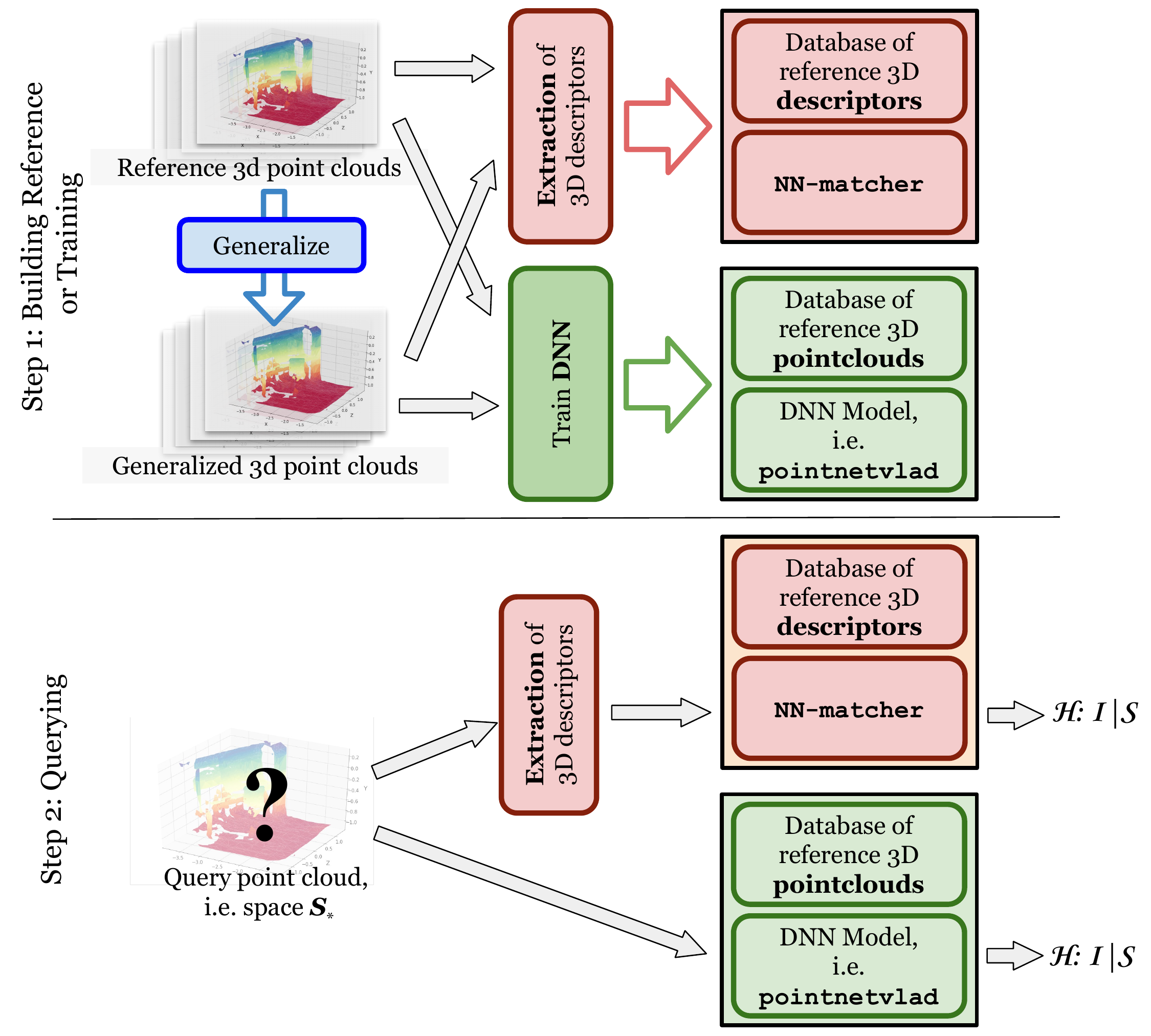}
	\vspace{-4mm}
	\caption{\small Reference construction, model training, and inference using two approaches: \texttt{NN-matcher}, and \texttt{pointnetvlad}}
	\label{fig:description_and_inference}
	\vspace{-4mm}
\end{figure}

We utilize two methods for our spatial inference attack: a descriptor nearest-neighbor matching approach we call \texttt{NN-matcher} (improved from \cite{deguzman2019firstlook}), and a deep neural network (DNN) based approach called \texttt{pointnetvlad} \cite{uy2018pointnetvlad}. 
Fig. \ref{fig:description_and_inference} shows a diagram of the overall process involved in the two inference methods. It has been previously demonstrated that spatial generalizations, i.e. 3D surfaces generalized as a set of planes, is an inadequate measure for spatial privacy \cite{deguzman2019firstlook}. Furthermore, generalizations can easily be replicated by an adversary allowing it to adjust its attack accordingly. Thus, as shown in Fig. \ref{fig:description_and_inference}, for both methods, we augment the raw captured point cloud data with multiple generalized versions as a combined reference ensemble. 

\begin{figure*}[t]
	\centering
	\vspace{-3mm}
	\begin{subfigure}{0.35\textwidth}
	    \centering
    	\includegraphics[width=\textwidth]{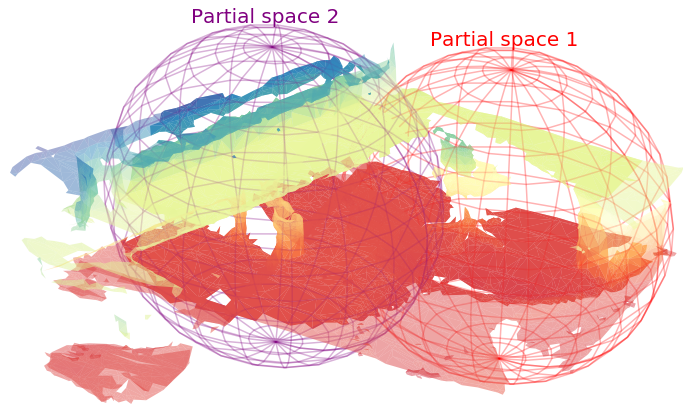}
    	\vspace{-2mm}
		\caption{\small Sample partial spaces of a bigger space}
    	\label{fig:partial-releases}
	\end{subfigure}
	\begin{subfigure}{0.295\textwidth}
	    \centering
    	\includegraphics[width=\textwidth]{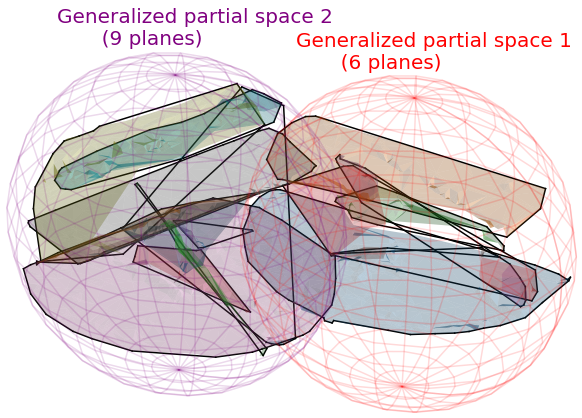}
    	\vspace{-2mm}
		\caption{\small Generalizing the partial spaces}
    	\label{fig:partial-generalized-releases}
	\end{subfigure}
	\begin{subfigure}{0.295\textwidth}
	    \centering
        	\includegraphics[width=\columnwidth]{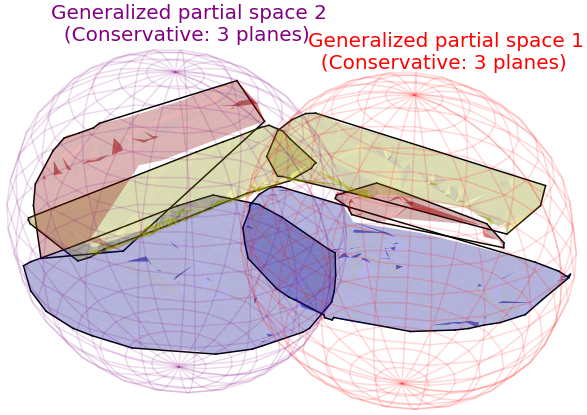}
    	\vspace{-2mm}
		\caption{\small Conservative release of planes}
    	\label{fig:partial-conservative-releases}
	\end{subfigure}
	\vspace{-2mm}
	\caption{\small Sample (a) partial releases with (b) generalization, and (c) conservative plane releasing.}
	\vspace{-2mm}
	\label{fig:protection-examples}
\end{figure*}

\subsection{Inference using 3d descriptors: \texttt{NN-matcher}}\label{subsec:NN-matcher-inference}

Step 1 for \texttt{NN-matcher} involves the construction of a reference set of descriptors from the historical data available augmented with generalized versions. A subset of oriented points for each space are selected and are called \textit{key points}. Then, for each key point, an \textit{intrinsic feature descriptor} is computed. The key point selection and feature computation depends on the chosen algorithm. The result of this process results to the accumulation of a set of key point-feature pairs for each reference space like so: $\footnotesize\{\{k_{i,0},f_{i,0}\},\{k_{i,1},f_{i,1}\}, ... \}$ for any space $i$. Then, for the inference step, we match the reference set with the key point-descriptor set of the unknown query space. 

For inter-space inference, we utilize a \textit{deterministic} matching-based approach using nearest-neighbor matching in the descriptor space, i.e. $\{f_{S^*}\} \mapsto \{f_{S_c \in \{{S_r}\}}\}$. Then, a \textit{voting mechanism} determines the best candidate match among the reference spaces ($S_c \in \{{S_r}\}$) which has the most matched descriptors with the query space ($S^*$). A similar voting-based mechanism was utilised in place recognition over 3D lidar data \cite{bosse2009keypoint,bosse2013place}.

We improve upon the mechanism used in \cite{deguzman2019firstlook} by extending the inference to intra-space location. We collect the key points of the corresponding features matched from inter-space inference, and trim the collection by only getting the pair of reference and query descriptors whose \textit{nearest neighbor distance ratio} (or NNDR) $< 0.9$. Then, using their corresponding key points, as in $\{k_{S^*}\} \mapsto \{k_{S_r}\}$,we perform a \textit{geometric structure consistency check} which produces the best pairs of key points with consistent structural relationships; that is, the graph (or sub-graph) of the reference key points is similar to the graph (or sub-graph) of the corresponding set of query key points. This can be generalized to the NP-complete sub-graph isomorphism problem but we instead use a heuristic-based approach using the product of the angular (using cosine similarity) and distance similarities of the two key point graphs. Then, we pick the resulting sub-graph with the most number of structurally consistent matched key points as the best intra-space match. We then compute the resulting intra-space distance $d(c_{S^*},c_{S})$ of the true centroid $c_{S}$ of the query space to that of the centroid of the matched reference space $c_{S^*} = c_{S^r}$ within the query space as described by Eq. \ref{eq:intra-space-distance}. 

\subsection{Inference using DNN: \texttt{pointnetvlad}}\label{subsec:pointnetvlad-inference}
To produce a large-scale place recognizer with 3D point cloud as input, \texttt{pointnetvlad} combines the deep network 3D point cloud shape classifier \texttt{PointNet} \cite{qi2017pointnet} with NetVLAD \cite{arandjelovic2016netvlad}, which is a deep network image-based place recognizer. To train \texttt{pointnetvlad}, we split the reference point clouds to disjoint training and validation sets. The point cloud sets are further subdivided to similarly-sized overlapping submaps with a radius of $2m$ and are re-sampled to contain 1024 points. The submap intervals are $0.5m$ and covers all 3 axes (while the original \texttt{pointnetvlad} only covers the two ground axes). The \texttt{PointNet} layer produces local feature descriptors for each point in a submap, and, then, feed those to the \texttt{NetVLAD} layer to learn a global descriptor vector for the given submap. We direct the reader to \cite{uy2018pointnetvlad} for more details on \texttt{pointnetvlad}.\footnote{Original \texttt{pointnetvlad} code can be accessed at \url{github.com/mikacuy/pointnetvlad}.}

For inference, \texttt{pointnetvlad} creates a reference database of the global descriptors of the combined raw and generalized spaces available. The point cloud of the query space will also be divided into submaps and be directly fed to the \texttt{pointnetvlad} model which likewise produces the two-level spatial inference hypothesis. 

\section{Privacy Measures over 3D Data}\label{sec:privacy-preservation}

Directly releasing raw point clouds exposes all spatial information as well as structural information of sensitive objects within the space. A mechanism can be inserted, as shown in Fig. \ref{fig:adversary-model-pipeline}, along the MR processing pipeline to provide privacy protection. We present two baseline protection measures: partial releasing, and planar spatial generalizations. However, it has been shown that planar spatial generalizations are inadequate forms of protection \cite{deguzman2019firstlook}; thus, we augment the protection further with \textit{conservative releasing} of the plane generalizations.


\subsection{Partial spaces}\label{subsec:partial-spaces} To limit the amount of information released with the point clouds, partial releasing can be utilised to provide MR applications the least information necessary to deliver the desired functionality. With partial spaces, we only release \textit{segments} of the space with varying radius. Fig. \ref{fig:partial-releases} shows an example space with 2 partial releases. Partial releasing can either be performed \textit{once} or up to a predefined number of releases if more of the space is necessary for the MR application to provide its service. Then, succeeding revelations of the space are no longer provided to the MR application. Moreover, partial releasing can be applied over raw or generalized spaces.

\subsection{Plane fitting generalization}\label{subsec:generalization}

As discussed in \S\ref{subsec:3d-data}, to deliver augmentations, MR platforms digitally maps the physical space to gain understanding of it. And, as discussed in \S\ref{subsec:mr-functionality}, depending on the desired functionality, an MR application may require just a single oriented point as a static anchor, a single plane or surface, or a set of surfaces for dynamic augmentations. Thus, arguably, without a significant impact on the delivery of the desired MR functionality, any surface within a space can be generalized into a set of planes. Furthermore, surface-to-plane generalizations inadvertently sanitizes information that is below the desired generalization resolution. For example, a keyboard on a desk surface may be generalized as part of the desk. However, spatial information, i.e. location, may still be inferred as we will reveal later. 

To perform surface-to-plane generalization, we employed the popular Random Sample Consensus (or RANSAC) plane fitting method \cite{fischler1981random}. 
For our implementation, we directly utilize the accompanying normal vector of each point to estimate the planes in the plane fitting process instead of computing or estimating them from the neighbouring points. 


\subsection{Conservative plane releasing}\label{subsec:conv-releasing} 
Plainly using plane generalizations do not adequately provide protection from spatial inference especially when we continuously reveal the space. Thus, we present \textit{conservative releasing}, where we limit the number of planes a generalization produces. Fig. \ref{fig:partial-generalized-releases} shows an example set of planes that are released after RANSAC generalization of the revealed partial raw spaces (in Fig. \ref{fig:partial-releases}); then, as shown in Fig. \ref{fig:partial-conservative-releases}, we can limit the maximum allowable planes that can be released to, say, a maximum of 3 planes in total.

As will be presented later in \S\ref{subsec:results-conservative}, conservative releasing is a viable privacy countermeasure which provides QoS that are arguably adequate for most MR functionalities.


\section{Evaluation Setup}\label{sec:methodology}

We present the various experimental setups we designed to investigate the viability of the different privacy-preserving methods described in \S\ref{sec:privacy-preservation}. We use the ensemble of raw point cloud data augmented with generalized versions as the reference set available to the adversary (Step 1 in Fig. \ref{fig:description_and_inference}). We, then, implement the various privacy-preserving methods to investigate how well can the adversary perform their attacks over such defenses. Specifically, for all succeeding experiments, we will be utilizing RANSAC generalization but with varying release mechanisms: in terms of the size of the partial space, the number of successive releases, and the maximum total number of released planes. 
We first present our data and specify our metrics before describing the evaluation setups. 

\subsection{Dataset}\label{subsec:dataset} 

\begin{figure}[t]
	\centering
	\vspace{-2mm}
	\includegraphics[width=\columnwidth]{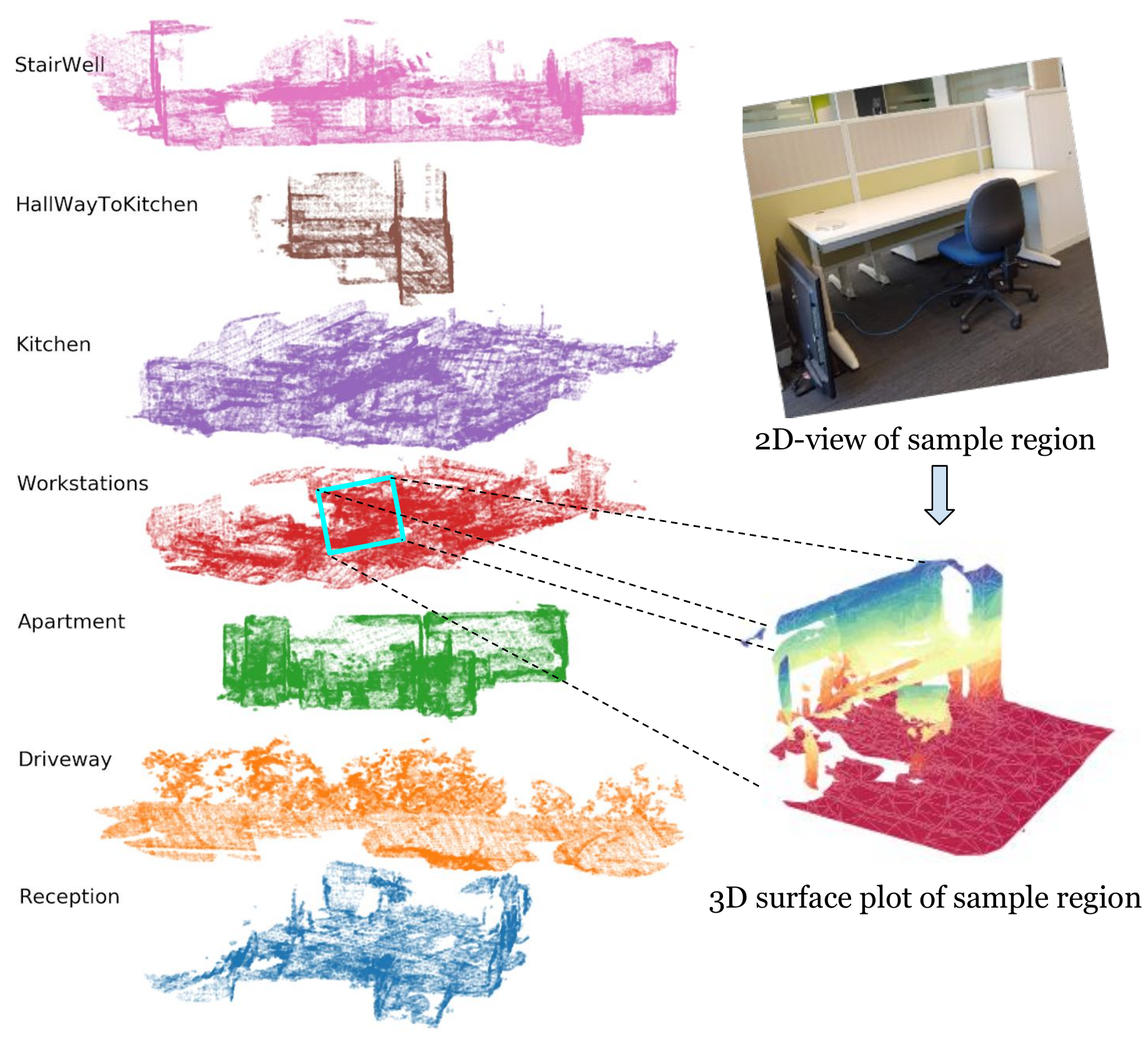}
	\vspace{-4mm}
	\caption{\small 3D point clouds of the 7 collected environments (left); a 3D surface of a sample space (bottom-right), and its 2D-RGB view (top-right).}
	\label{fig:point-cloud}
	\vspace{-5mm}
\end{figure}

For our dataset, we gathered real 3D point cloud data using the Microsoft HoloLens in various environments to demonstrate the leakage from actual human-scale spaces in which an MR device is usually used. 
As shown in Fig. \ref{fig:point-cloud}, our collected environments include the following spaces: a work space, a reception area, an office kitchen or pantry, an apartment, a drive way, a hall way, and a stair well.

The combined approximate floor area of the spaces is $899.9{m^2}$ while the combined approximate surface area (i.e. including the vertical surfaces and other objects whose 3d point cloud are captured) is $1434.7{m^2}$. The combined size of the raw point clouds are $39.7$MB. The resulting reference descriptor set used by \texttt{NN-matcher} is $5.5$MB while the reference database used by \texttt{pointnetvlad} is $5.41$MB.

\subsection{Metrics}\label{subsec:metrics} To quantify the privacy leakage in our various evaluation setups, as defined in \S\ref{subsec:adversary-model}, we use adversarial inference error as our privacy metrics. We posed a two-level attack: (1) inter-space inference, and (2) intra-space inference. For the first level, privacy can be directly linked, as in Eq. \ref{eq:inter-space-privacy}, to the inter-space inference error rate. While, for the second level, as in Eq. \ref{eq:intra-space-privacy}, intra-space privacy can be related to the distance error which is in distance units $u$ (where $1~u$ is approximately $1~meter$).

Moreover, we set the following desired subjective lower-bounds for the privacy metrics. For inter-space privacy, we can set a desirable lower-bound at $\Pi_{1}\geq0.5$. This means that an adversary can only make a correct guess at most half the time. Furthermore, we define $1.0\geq\Pi_{1}\geq0.75$ as high privacy, $0.75>\Pi_{1}\geq0.5$ as medium privacy, and  $\Pi_{1}<0.5$ as low privacy. For intra-space privacy, we set a desirable lower-bound at $\Pi_{2}\geq4m$. However, we emphasize the great subjectivity of these lower-bounds especially that of the intra-space distance error where a desirable lower-bound highly depends on the scenario or location. For example, for indoor scenarios, a distance error of at least $4m$ can perhaps mean that the actual user is in a different room, while, for outdoor scenarios, a distance error of at least $4m$ is still relatively small.

\subsection{Setup}\label{subsec:partial-releasing}

We validate the viability of generalization as we vary the size of the one-time partially released generalized spaces, and as we successively release more partial spaces. Then, we proceed with the investigation of the viability of conservative releasing as an augmented countermeasure to generalizations. 


%
\paragraph{One time release of partial spaces.}\label{subsubsec:one-time-release} For initial investigation and validation, we use the case when an MR application is provided \textit{only once} with 3D spatial information but we vary the size, i.e. radius $r$, of the revealed space. We apply this to the raw point clouds as well as for RANSAC-generalized point clouds. For every radius $r$, we get 1000 random sample spaces, i.e. random submap of radius  $r$ from a randomly picked space, as a user can initiate an MR application from any point within a space; to ensure rotation-invariance, we also randomly rotate and translate the space. We, then, measure the inference performance. We feed the same set of partial spaces to the two chosen attack approaches: the descriptor-based \texttt{NN-matcher}, and the deep neural network-based \texttt{pointnetvlad}.

\paragraph{Successive release of partial spaces.}\label{subsubsec:successive-release}
To demonstrate the case when users are moving around and their physical space is gradually revealed, we included a validation setup that successively releases partial spaces. 
Following the described generalization strategy in \S\ref{subsec:generalization}, we perform successive releasing of partial spaces for collected raw point cloud, and for RANSAC generalized versions. We do 100 random sample partial iterations, and produce 100 releases per random sample. We do this for radii $= \{ 0.5, 1.0, 2.0\}$.

As the released points are accumulated, we perform generalization over the accumulated points. 
To achieve consistency on the generalizations, we implemented a \textit{plane subsumption} handler for generalizing successively released points. Specifically, new points will be checked against existing planes (produced by previous releases) if they can be subsumed by the existing ones instead of performing RANSAC generalization for the entire accumulated points; RANSAC will only be performed over the remaining [ungeneralized] points. 

We feed the same set of successive spaces to both the \texttt{NN-matcher}, and \texttt{pointnetvlad}. The resulting stronger attacker will be used for the subsequent cases.

\paragraph{Conservative plane releasing.}\label{subsubsec:conv-releasing} In addition to the previous setups, we employ \textit{conservative releasing} which limits the number of planes a surface-to-plane generalization produces as a form of spatial privacy countermeasure. For our investigation, we apply conservative releasing over the same set of successively released partial spaces with subsumption applied during RANSAC generalization. But, for every sample release, we investigate the impact of limiting the maximum number of planes, we label $\mathtt{max\_planes}$, that are generated by the RANSAC plane generalization process. We performed controlled plane releasing with $\mathtt{max\_planes}$ in steps of 2 from 1 to 30. 

\section{Results}\label{sec:results}

Now, we present the results of the systematic evaluation. We first present the results of the validation experiments over partially released and successively released spaces followed by the results of our investigation on conservative releasing as a complementary approach to generalization for privacy preservation.

\subsection{One time partial releasing}\label{subsec:results-partial}

\begin{figure}[t]
	\vspace{-3mm}
	\includegraphics[width=\columnwidth]{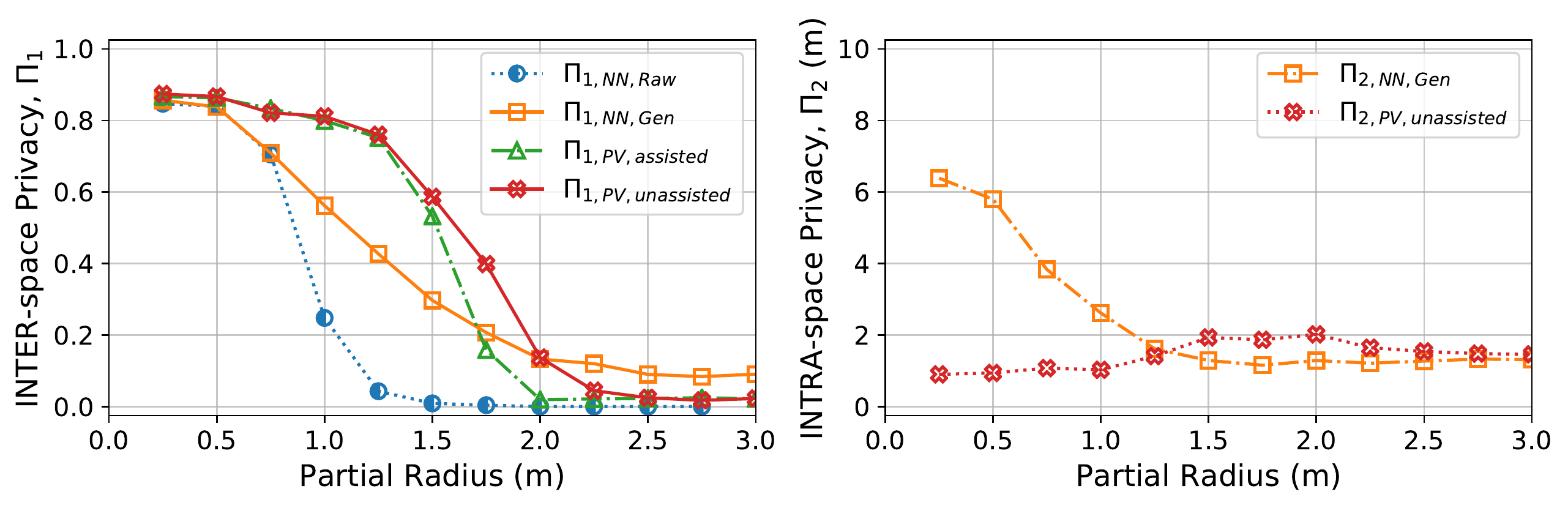}
	\vspace{-5mm}
	\caption{\small One-time partially released RANSAC-generalized spaces vs varying radii: (left) inter-space and (right) intra-space privacy}
	\label{fig:partial-radius}
	\vspace{-5mm}
\end{figure}

Fig. \ref{fig:partial-radius} shows the average privacy provided by surface planar generalization as we vary the size of released spaces for the one-time release case, and for our two attackers. We use the subscript \texttt{\textit{NN}} for \texttt{NN-matcher} while \texttt{\textit{PV}} for \texttt{pointnetvlad}. We also show the privacy values of Raw spaces for comparison. As shown in Fig. \ref{fig:partial-radius}-left, generalization provides improved inter-space privacy for the one-time partial release case. This can be observed in the sharp drop of $\Pi_{\texttt{1,NN,Raw}}$ while $\Pi_\texttt{1,NN,Gen}$ slowly drop as we increase the radius. At $r = 1.0m$, there is more than a two-fold difference between $\Pi_\texttt{1,NN,Raw}$ and $\Pi_\texttt{1,NN,Gen}$; at $r = 1.5m$, $\Pi_\texttt{1,NN,Raw}$ is already $<0.1$ while $\Pi_\texttt{1,NN,Gen}$ is still $>0.3$.

On the other hand, \texttt{pointnetvlad} learns to generalize during training, thus raw and surface-generalized query spaces will result to the similar global descriptors and, hence, same $\Pi_1$ values. As shown in Fig. \ref{fig:partial-radius}-left, \texttt{pointnetvlad} performs worse than the \texttt{NN-matcher} at $r<1.75m$. This is due to the submaps used to train \texttt{pointnetvlad} having $2m$ radius.\footnote{In the interest of space, we no longer show the results of the preliminary analysis over \texttt{pointnetvlad} but smaller submaps have worse performance for two primary reasons: smaller submaps (1) contain less information, and (2) results to more similar-looking submaps.} To further demonstrate the impact of submap generation, we show the performance in Fig. \ref{fig:partial-radius}-left when we \textit{assist} the submap generation with true centroids of the partial spaces, and when it is unassisted--i.e. the attacker has to infer the centroids of the partial spaces. When assisted, a slight improvement on \texttt{pointnetvlad}'s performance can be observed: i.e. at $r=2.0m$, $P_\texttt{1,PV,assisted} < 0.1$ while $P_\texttt{1,PV,unassisted} > 0.1$; however, this only further exposes the weakness of \texttt{pointnetvlad} for $r<2.0m$. In the subsequent cases, we only show the results of unassisted \texttt{pointnetvlad}.

Subsequently, Fig. \ref{fig:partial-radius}-right shows the average intra-space privacy for the partial query spaces that are \textit{correctly labelled} during inter-space inference. We only focus on the results for the surface-generalized query spaces, and only for the unassisted \texttt{pointnetvlad}. At partial releases with $r\leq0.75m$, $\Pi_\texttt{2,NN} > 4m$ which directly translates to the intra-space hypothesis being off by at least $4m$ in average from the true intra-space location. On the other hand, $\Pi_\texttt{2,PV} \leq 2m$ regardless of the radius: the \texttt{NetVLAD} layer ensures that nearby submaps will have similar global descriptors (and distant submaps to be dissimilar) which leads to good intra-space performance. On the other hand, \texttt{NN-matcher} descriptors are directly computed from the point cloud which means that nearby descriptors can be dissimilar while distant descriptors can be similar.


Evidently, there is no intra-space privacy benefit from generalizations for $r>1.0m$, but, as shown in Fig. \ref{fig:partial-radius}, a significant inter-space privacy benefit can be observed. 
However, one-time partial releasing provides very limited data to applications. In reality, other MR applications may desire to receive new, expanded, and/or updated information about the user's physical space to deliver their MR functionality with immersiveness. Thus, we also need to investigate the viability of using generalizations but with successively released spatial information. 

\vspace{-5mm}\textit{\paragraph{\textbf{Takeaway}.} Surface-to-plane generalizations provide INTER-space privacy benefit for the one-time partial release case, but does not provide any significant benefit in terms of INTRA-space privacy. 
}


\subsection{Successive releasing}\label{subsec:results-successive}

\begin{figure}[t]
	\vspace{-3mm}
	\includegraphics[width=\columnwidth]{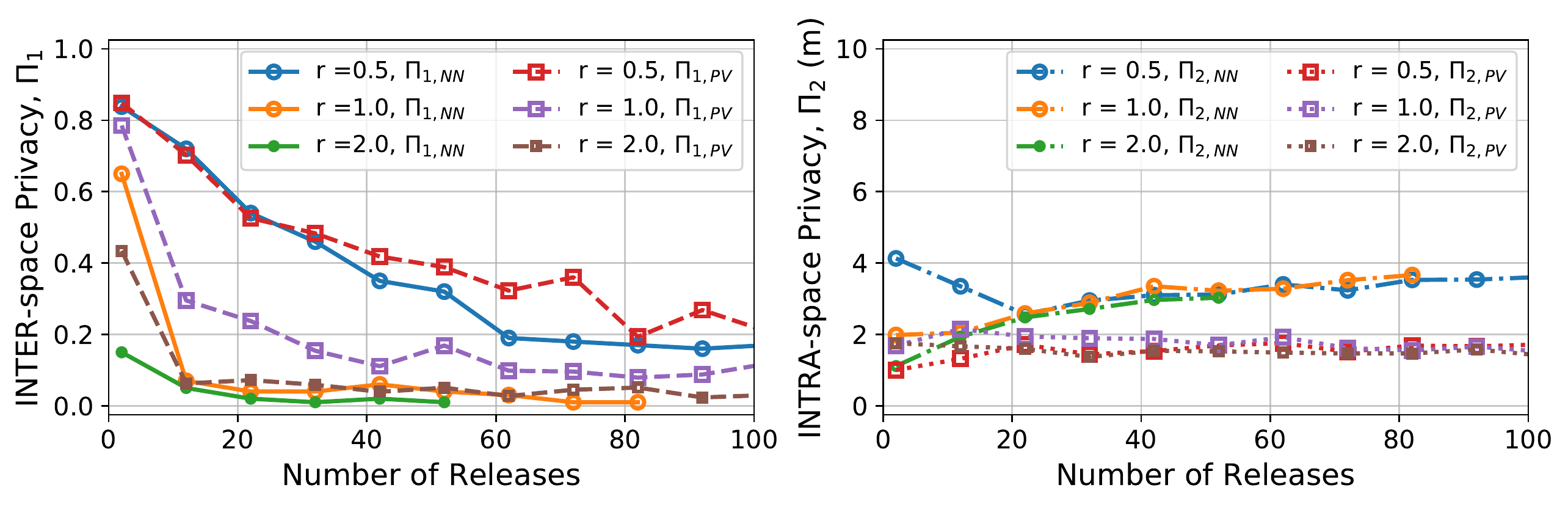}
	\vspace{-4mm}
	\caption{\small Successively released partial spaces: (left) inter-space and (right) intra-space privacy}
	\label{fig:successive-all-ransac}
	\vspace{-4mm}
\end{figure}

Fig. \ref{fig:successive-all-ransac} shows the inference performance as we successively release generalized partial spaces. Regardless of the size of the released partial spaces and of the attacker used, as we either increase the size of a partial space, or reveal more portions and/or planes of the space, unsurprisingly, the inter-space privacy (Fig. \ref{fig:successive-all-ransac}-left) decreases. 
For $r = 0.5m$, both $\Pi_\texttt{1,NN}$ and $\Pi_\texttt{1,PV}$ slowly drops but with $\Pi_\texttt{1,NN}$ dropping faster. If we double the size of the releases to $r = 1.0m$, both $\Pi_1$ drops quickly with $\Pi_\texttt{1,NN}$ still dropping faster. Thus, in terms of inter-space privacy, \texttt{NN-matcher} still performs better than the deep neural network-based \texttt{pointnetvlad} even in the successive case.

Despite the accumulated spaces in the successive case being larger than $2.0m$ after a few releases, \texttt{pointnetvlad}'s performance still suffers from non-optimal submap generation. We did observe that $\Pi_1$ drops to $0$ in Fig. \ref{fig:partial-radius}-left, but that is due to how the partial spaces are generally enclosed in spheres. Whereas in the successive case, the accumulated spaces will now be irregularly shaped. Fig. \ref{fig:protection-examples} shows how it looks with two releases. Thus, for accumulated spaces larger than $2.0m$, the submap generator will produce more than 1 submap to cover the entire query space and, then, perform the inference on the submaps. Some of these submaps, say the edge submaps, will lead to erroneous inter-space labels. On the other hand, similar to the one-time partial case, intra-space performance of \texttt{pointnetvlad} is fairly sustained and better than the \texttt{NN-matcher}--owing to the discriminative power of the \texttt{NetVLAD} layer as long as the inter-space label is correct.

Differently, the \texttt{NN-matcher}'s intra-space distance error initially drops but slowly increases as we reveal more of the space. 
As shown in Fig. \ref{fig:successive-all-ransac}-right, at the first release, at $r = 0.5m$, $\Pi_\texttt{2,NN}$ is slightly high with an error of $\geq4m$, but, after 20 or more releases, the  $\Pi_{2}$ drops $<3m$. But, interestingly, as we reveal more of the space, the $\Pi_2$ seems to improve: for $r = 1.0m$, $\Pi_{2}$ approximately increases to almost $4m$ again as we reach 60 or more releases. 
And, regardless of the size, after 20 releases, $\Pi_\texttt{2,NN}$ follows a similar increasing trend. 
This can be attributed to how artificial, i.e. man-made, spaces have repeating similar structures which also produces very similar descriptors. As a result, successive releases can contain structures on a different intra-space location but is similar to structures that were previously released. A good example of this are the workstations in the same company or institution which are all designed similarly.

However, this seemingly improving intra-space privacy is not necessarily an improvement as the revealed space grows in size which leads to a possible overlap of the hypothesis intra-space to the true space. For example, after a number of releases the width of the accumulated space is $4.0m$, a $\Pi_\texttt{2,NN}<4.0m$ means that the adversary was still able to produce a hypothesis space that overlaps with or, even, within the true space. 
Nevertheless, we will use \texttt{NN-matcher} in the succeeding setups as it outperforms \texttt{pointnetvlad} in inter-space inference.

\vspace{-5mm}\textit{\paragraph{\textbf{Takeaway}.} The evaluation over successive releases highlights the inference power, especially in INTER-space inference, of a descriptor matching-based attacker in recognizing spatial locations from 3D MR data. 
}

\subsection{Conservative releasing}\label{subsec:results-conservative}

\begin{figure}[t]
	\vspace{-3mm}
    \begin{subfigure}{0.44\columnwidth}
		\centering
			\includegraphics[width=\textwidth]{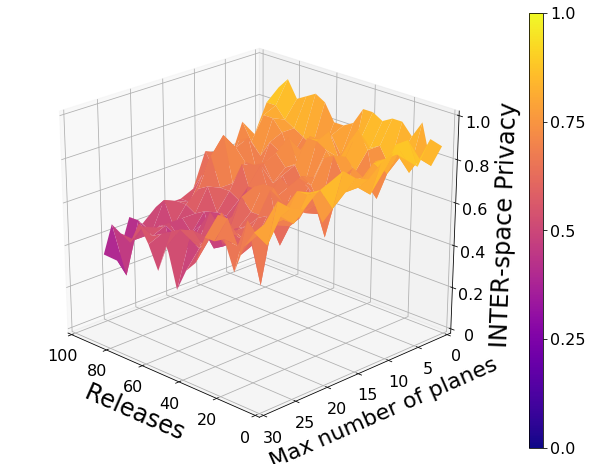}
		\caption{\small 3D plot ($r = 0.5$)}
			\label{fig:varying-planes-3D}
	\end{subfigure}
	\hspace{4mm}
    \begin{subfigure}[]{0.48\columnwidth}
		\centering
			\includegraphics[width=\textwidth]{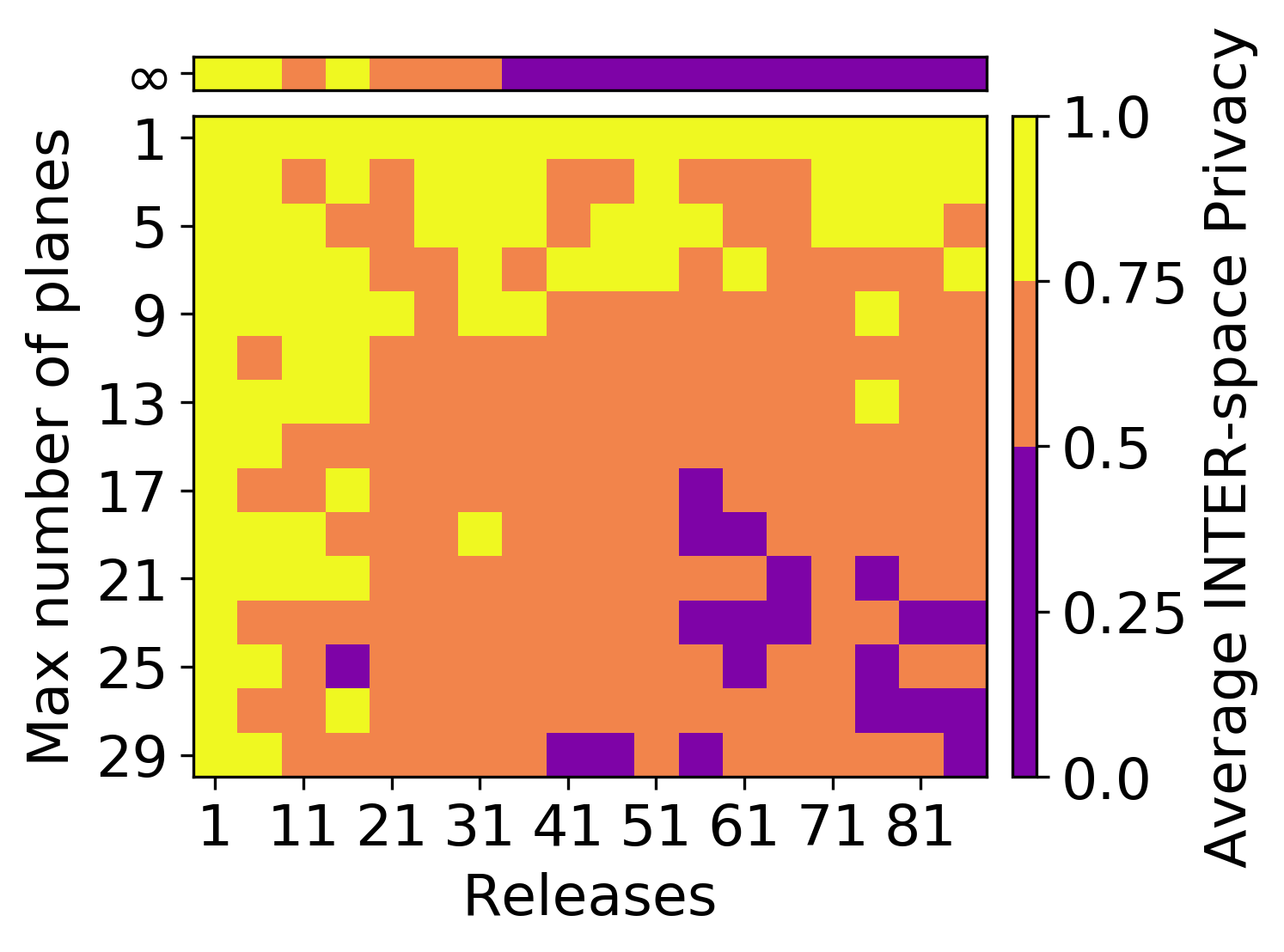}
		\caption{\small Heatmap ($r = 0.5$)}
			\label{fig:varying-planes-heatmap}
	\end{subfigure}
    \begin{subfigure}[]{0.44\columnwidth}
	\vspace{2mm}
		\centering
			\includegraphics[width=\textwidth]{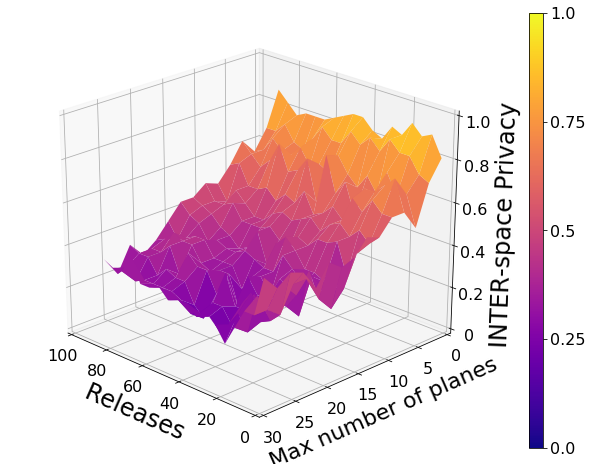}
		\caption{\small Heatmap ($r = 1.0$)}
			\label{fig:varying-planes-3D-r1}
	\end{subfigure}
	\hspace{4mm}
    \begin{subfigure}[]{0.48\columnwidth}
	\vspace{2mm}
		\centering
			\includegraphics[width=\textwidth]{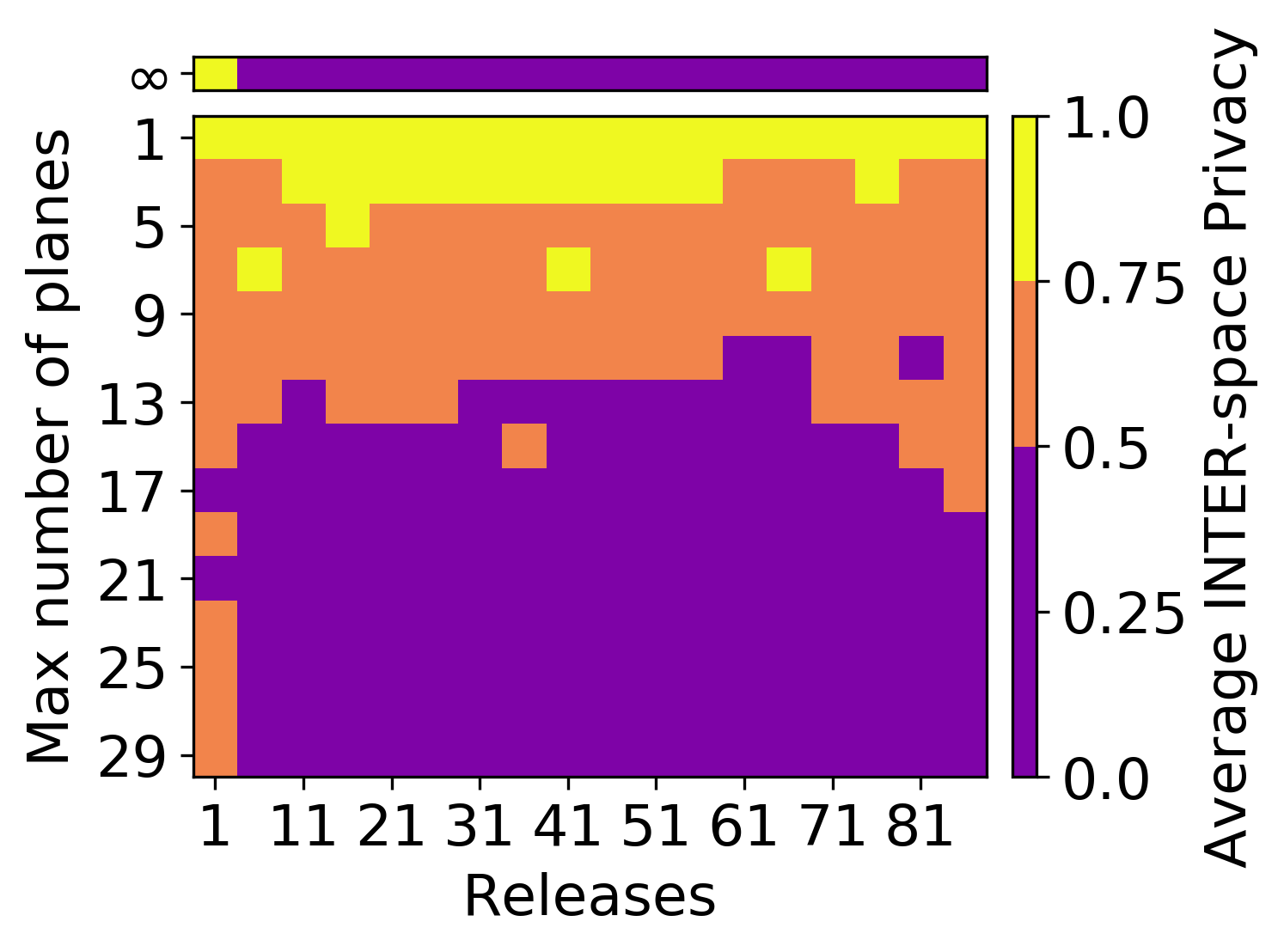}
		\caption{\small Heatmap ($r = 1.0$)}
			\label{fig:varying-planes-heatmap-r1}
	\end{subfigure}
    \begin{subfigure}[]{\columnwidth}
        \vspace{2mm}
		\centering
			\includegraphics[width=\textwidth]{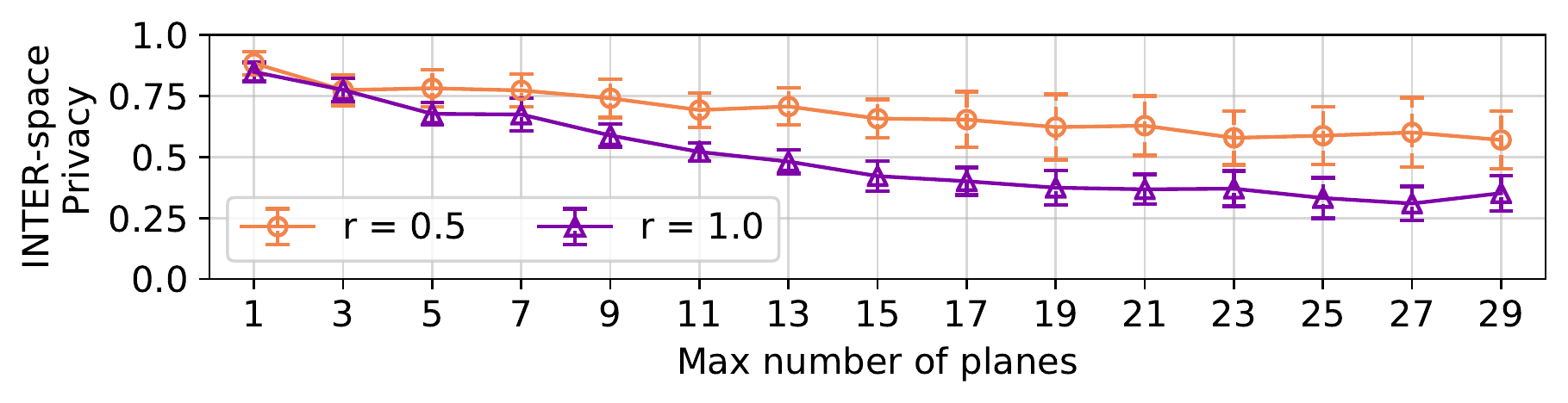}
		\caption{\small Average $\Pi_1$ over all releases}
			\label{fig:varying-planes}
	\end{subfigure}
	\vspace{-4mm}
	\caption{\small Average INTER-space privacy of conservatively released planes over successive releasing (using \texttt{NN-matcher} attacker)}
	\label{fig:conservative-releasing-inter}
	\vspace{-4mm}
\end{figure}

\begin{figure}[t]
	\vspace{-3mm}
    \begin{subfigure}[]{0.44\columnwidth}
		\centering
			\includegraphics[width=\textwidth]{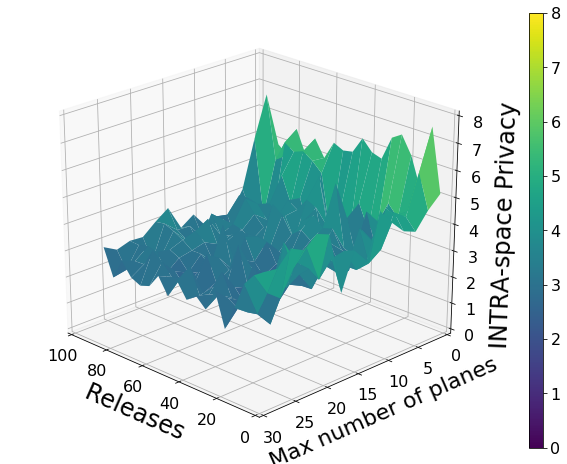}
		\caption{\small 3D plot ($r = 0.5$)}
			\label{fig:varying-planes-3D-intra}
	\end{subfigure}
	\hspace{4mm}
    \begin{subfigure}[]{0.48\columnwidth}
		\centering
			\includegraphics[width=\textwidth]{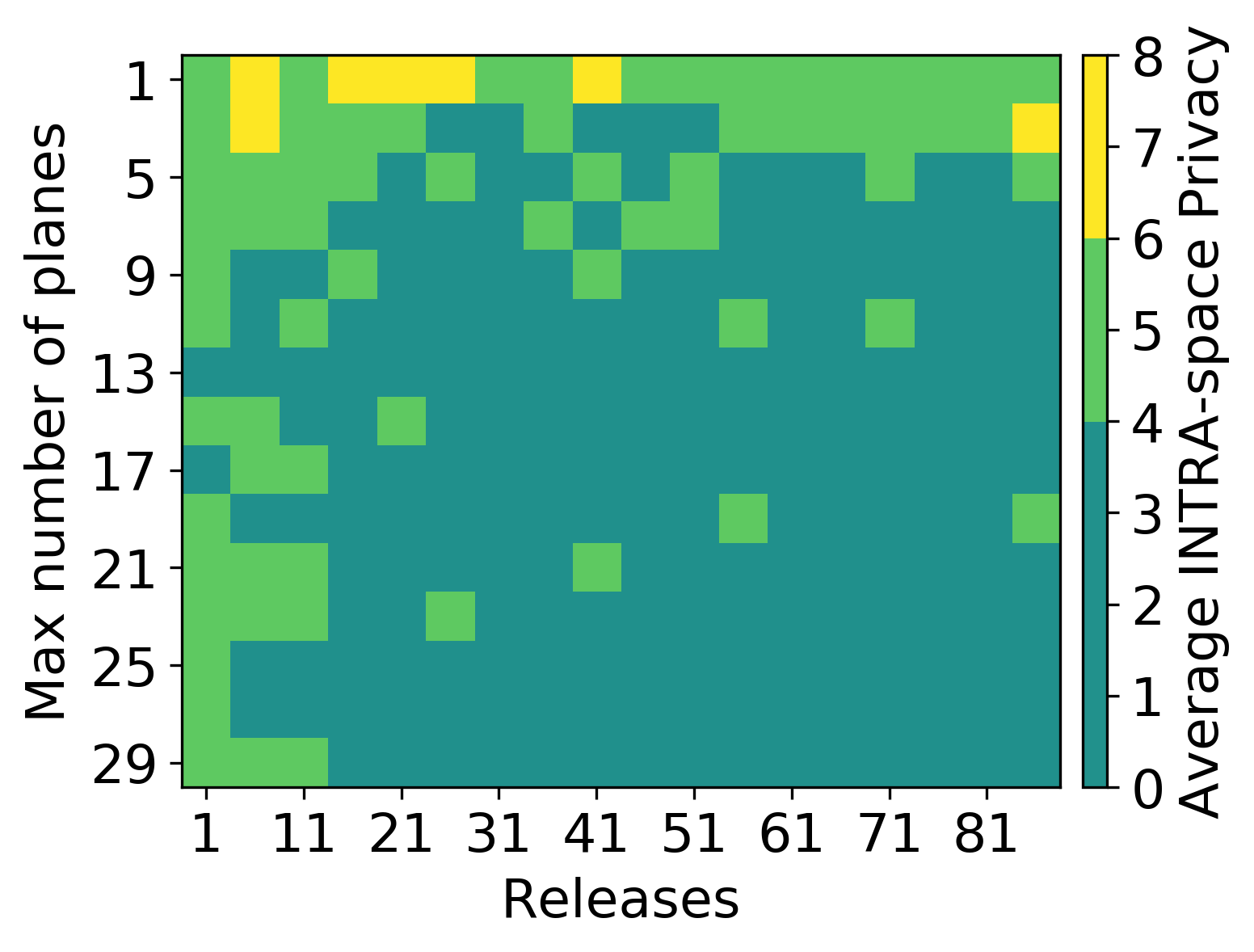}
		\caption{\small Heatmap ($r = 0.5$)}
			\label{fig:varying-planes-heatmap-intra}
	\end{subfigure}
    \begin{subfigure}[]{0.44\columnwidth}
	\vspace{2mm}
		\centering
			\includegraphics[width=\textwidth]{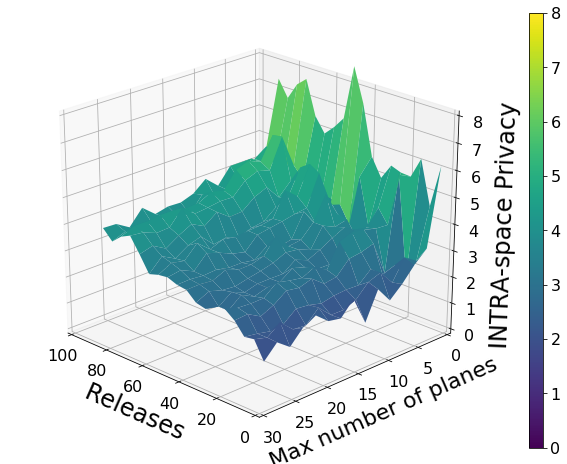}
		\caption{\small 3D plot ($r = 1.0$)}
			\label{fig:varying-planes-3D-intra-r1}
	\end{subfigure}
	\hspace{4mm}
    \begin{subfigure}[]{0.48\columnwidth}
	\vspace{2mm}
		\centering
			\includegraphics[width=\textwidth]{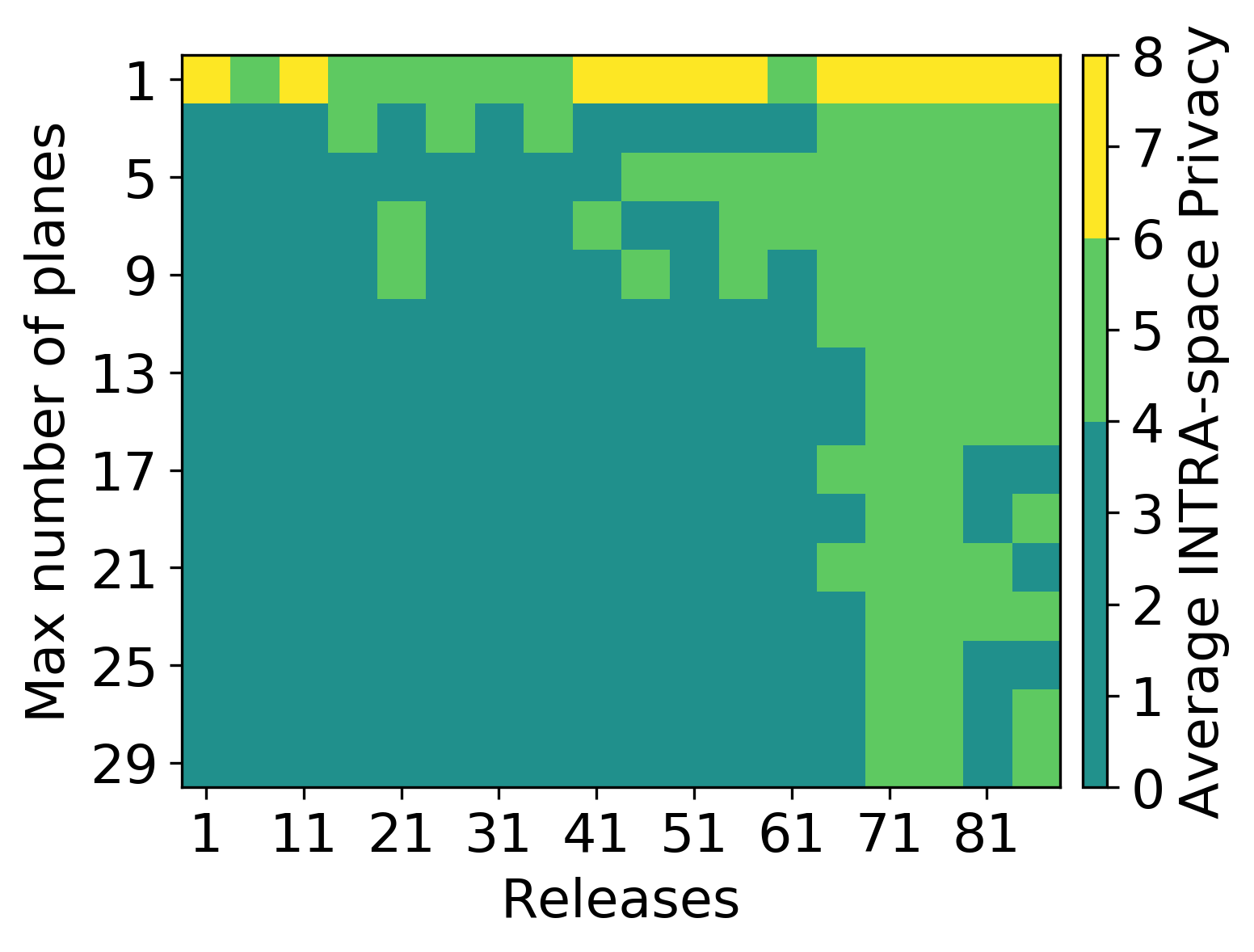}
		\caption{\small Heatmap ($r = 1.0$)]}
			\label{fig:varying-planes-heatmap-intra-r1}
	\end{subfigure}
    \begin{subfigure}[]{\columnwidth}
	\vspace{2mm}
		\centering
			\includegraphics[width=0.98\textwidth]{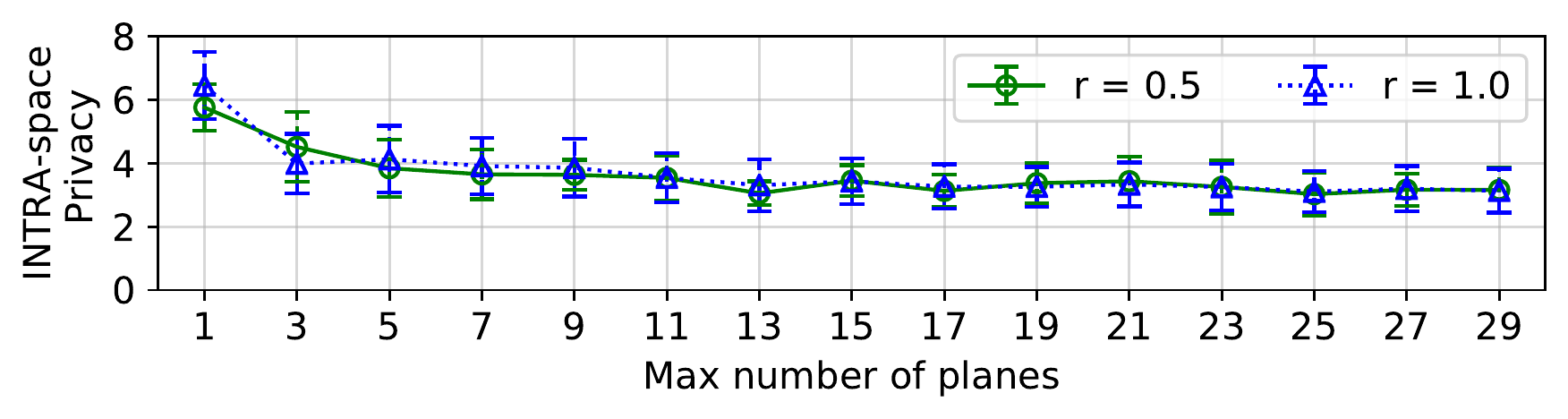}
		\caption{\small Average $\Pi_2$ over all releases}
			\label{fig:varying-planes-intra}
	\end{subfigure}
    \begin{subfigure}[]{\columnwidth}
	\vspace{2mm}
		\centering
			\includegraphics[width=0.98\textwidth]{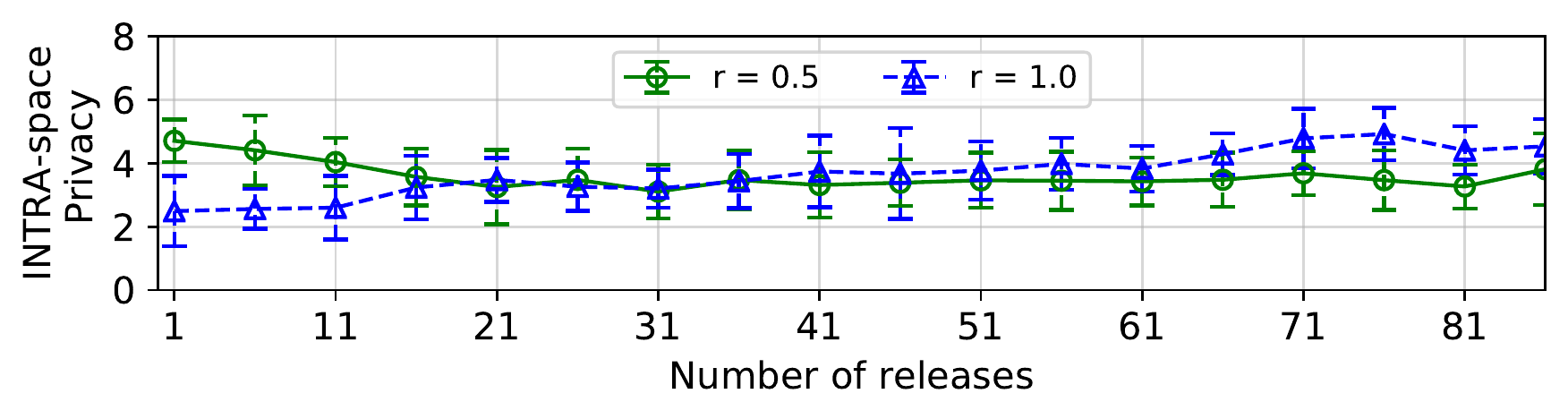}
		\caption{\small Average $\Pi_2$ over varying number of planes}
			\label{fig:varying-planes-intra-releases}
	\end{subfigure}
    \vspace{-4mm}
	\caption{\small Average INTRA-space privacy of conservatively released planes over successive releasing (using \texttt{NN-matcher} attacker)}
	\label{fig:conservative-releasing-intra}
	\vspace{-4mm}
\end{figure}

Now, we explore the privacy approach of combining \textit{conservative releasing} with surface planar generalizations to provide spatial privacy in MR as more of the space is revealed. Specifically, we control the maximum number of released planes after the surface-to-plane generalization process. 
Fig. \ref{fig:conservative-releasing-inter} shows the inter-space privacy values as we increase the maximum number of planes, i.e. $\mathtt{max\_planes}$, and the number of releases. We can see from the 3D plot for $r = 0.5m$ in Fig. \ref{fig:varying-planes-3D} that the inter-space privacy gradually decreases as we reveal more of the space (more releases) and increase $\mathtt{max\_planes}$. For $\Pi_{1}(r = 0.5m)$ to go below $0.5$, both the number of releases and $\mathtt{max\_planes}$ needs to be high. This is more evident through the unevenly quantized heatmap shown in Fig. \ref{fig:varying-planes-heatmap} which shows that we will get $\Pi_{1} \leq0.5$ for only a few occasions in the provided range of values for number of releases and $\mathtt{max\_planes}$.\footnote{The uneven quantized levels follows our set ranges in \S\ref{subsec:metrics}} We also show the heatmap version for the successive case with no conservative releasing, i.e. $\mathtt{max\_planes}=\infty$, at the top of Fig. \ref{fig:varying-planes-heatmap} for comparison. And as shown, for successively released partial spaces with radius $r = 0.5m$, the average inter-space privacy $\Pi_{1}$ drops below $0.5$ after 31 or more releases. But, with conservative releasing, we can release up to $51$ successive partial releases with $\mathtt{max\_planes}\leq 23$ for $\Pi_{1} \geq0.5$. Furthermore, if we average $\Pi_{1}(r = 0.5m)$ over all releases and observe it relative to $\mathtt{max\_planes}$, average $\Pi_{1}(r = 0.5m)$ does not go below $0.5$ if $\mathtt{max\_planes}\leq 29$ as shown in Fig. \ref{fig:varying-planes}.

On the other hand, for $r = 1.0m$, Figs. \ref{fig:varying-planes-3D-r1} and \ref{fig:varying-planes-heatmap-r1} shows that the decrease in $\Pi_{1}$ is primarily due to $\mathtt{max\_planes}$. Aside from the few occasions shown in Fig. \ref{fig:varying-planes-heatmap-r1} that $\Pi_1 \geq 0.5$ with lower number of releases, we can observe that generally $\Pi_1 < 0.5$ for $\mathtt{max\_planes}\geq13$. We also show the successive case with $\mathtt{max\_planes}=\infty$ for $r = 1.0m$ which shows that right after the first release we have $\Pi_1 < 0.5$. Furthermore, as shown in Fig. \ref{fig:varying-planes}, the average $\Pi_{1}(r = 1.0m)$ over all releases goes below $0.5$ if the $\mathtt{max\_planes}\geq 13$. Thus, for $r \leq 1.0$, there can be at most 11 planes released regardless of the number of successive partial releases so that $\Pi_1(r = 1.0m) \geq 0.5$. Arguably, having a maximum limit of 11 $planes$ is already adequate for most MR functionalities.

We also present the intra-space privacy with conservative releasing in Fig. \ref{fig:conservative-releasing-intra} with quantized heatmaps. 
Figs. \ref{fig:varying-planes-3D-intra}-\ref{fig:varying-planes-heatmap-intra} and  \ref{fig:varying-planes-3D-intra-r1}-\ref{fig:varying-planes-heatmap-intra-r1} shows differences in $\Pi_2$ for $r = 0.5$ and $1.0m$, but, if we are to average it over all releases as shown in Fig. \ref{fig:varying-planes-intra}, the plots of $\Pi_2(r = 0.5m)$ and $\Pi_2(r = 1.0m)$ overlap and plateaus at roughly $\sim 3.085m$. The difference being only at $\mathtt{max\_planes} \leq 3$ where $\Pi_2(r = 1.0m)$ drops faster $\Pi_2(r = 0.5m)$ from $\mathtt{max\_planes} = 1$ to $3$ before eventually plateauing at similar levels. 

On the other hand, if we are to average $\Pi_2$ over $\mathtt{max\_planes}$, we see that Fig. \ref{fig:varying-planes-intra-releases} reflects the behavior of $\Pi_\texttt{2,NN}$ shown in Fig. \ref{fig:successive-all-ransac}-right. Both figures show increasing $\Pi_2$ as we increase the number of releases. In Fig. \ref{fig:varying-planes-intra-releases}, it is more evident for $\Pi_2(r = 1.0)$ which starts to increase immediately after 1 release, while $\Pi_2(r = 0.5)$ increases only after 31 releases.

\vspace{-5mm}\textit{\paragraph{\textbf{Takeaway}.} Conservative releasing of generalized planes provide INTER-space privacy: for $r = 0.5$, if we are to release up to $17$ planes, an adversary will misidentify a space at least half of the time regardless of the number of releases, and will be off by at least $3.0m$ in INTRA-space inference for these infrequent occasions of INTER-space success. We can get similar privacy for $r = 1.0$ with up to $11$ plane releases.}

\subsection{Utility in terms of QoS}\label{subsec:utility-results}

\begin{figure}[t]
	\vspace{-3mm}
	\centering
	\includegraphics[width=0.85\columnwidth]{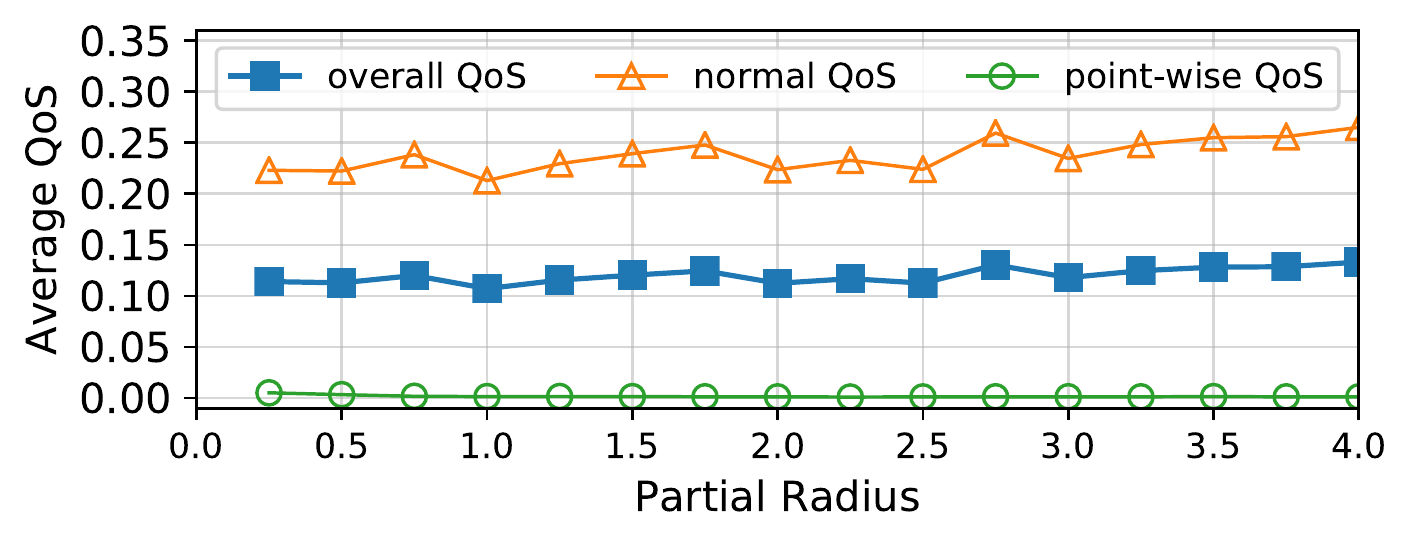}
	\vspace{-4mm}
	\caption{\small QoS $Q$ vs varying radius $r$. The average $Q$ over all test radii, i.e. $0.25\leq r\leq4.0m$, is $0.120\pm0.007$, while there is a \textbf{medium} positive correlation, i.e. $\rho = 0.71$, between $Q$ and $r$.}
	\label{fig:partial-radius-utility}
	\vspace{-4mm}
\end{figure}

Plane-fitting generalizations contribute variations to the released point clouds from true spaces. Fig. \ref{fig:partial-radius-utility} shows the computed average QoS $Q$ based on Eq. \ref{eq:utility-by-qos} (with coefficients $\alpha,\beta = 0.5$) for RANSAC generalized spaces with varying radii $r$. 
We also show the plots of two components used to compute the overall $Q$: the normal QoS, and the point-wise QoS.\footnote{A normal QoS of $0.2$ means that the resulting normals after generalization are on average about $18^\circ$ off from the original normals.} As shown, the overall $Q$ very slowly increases (but not consistently) as we increase the radius. The increase can be attributed more to the normal QoS component which has medium positive correlation, i.e. $\rho = 0.74$, with radius $r$. While the overall $Q$ has a medium positive correlation, i.e. $\rho = 0.71$, with $r$, and, if we average $Q$ over all radii, we get $\Bar{Q} = 0.120\pm0.007$ which has less than $10\%$ standard deviation. This slight increase is attributed to how more information, i.e. larger radius, introduces additional errors, albeit minimally, especially on the normals during generalization.

\begin{figure}[t]
 	\vspace{-2mm}
   \begin{subfigure}[]{0.44\columnwidth}
		\centering
			\includegraphics[width=\columnwidth]{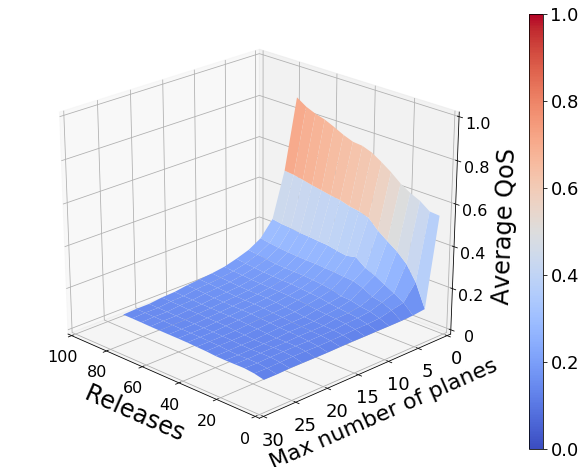}
		\caption{\small 3D plot ($r = 0.5$)}
			\label{fig:varying-planes-3D-utility}
	\end{subfigure}
	\hspace{4mm}
    \begin{subfigure}[]{0.48\columnwidth}
		\centering
			\includegraphics[width=\columnwidth]{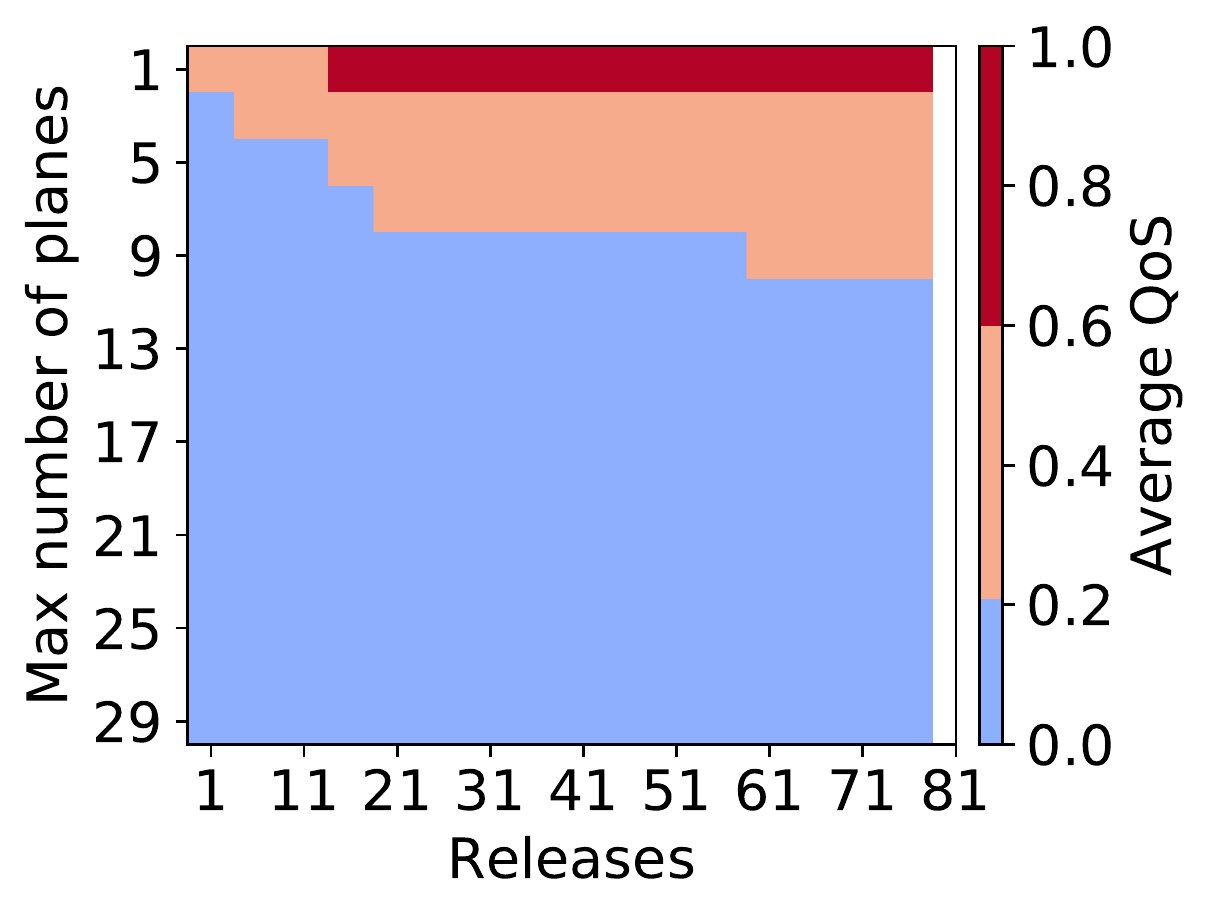}
		\caption{\small Heatmap ($r = 0.5$)}
			\label{fig:varying-planes-heatmap-utility}
	\end{subfigure}
    \begin{subfigure}[]{0.44\columnwidth}
		\vspace{2mm}
		\centering
			\includegraphics[width=\columnwidth]{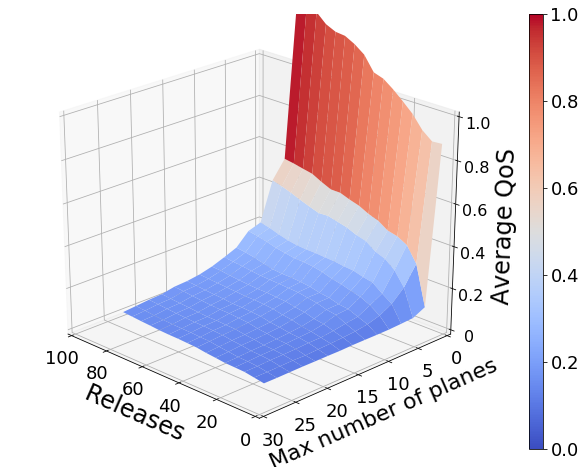}
		\caption{\small 3D plot ($r = 1.0$)}
			\label{fig:varying-planes-3D-utility-r1}
	\end{subfigure}
	\hspace{4mm}
    \begin{subfigure}[]{0.48\columnwidth}
        \vspace{2mm}
        \centering
			\includegraphics[width=\columnwidth]{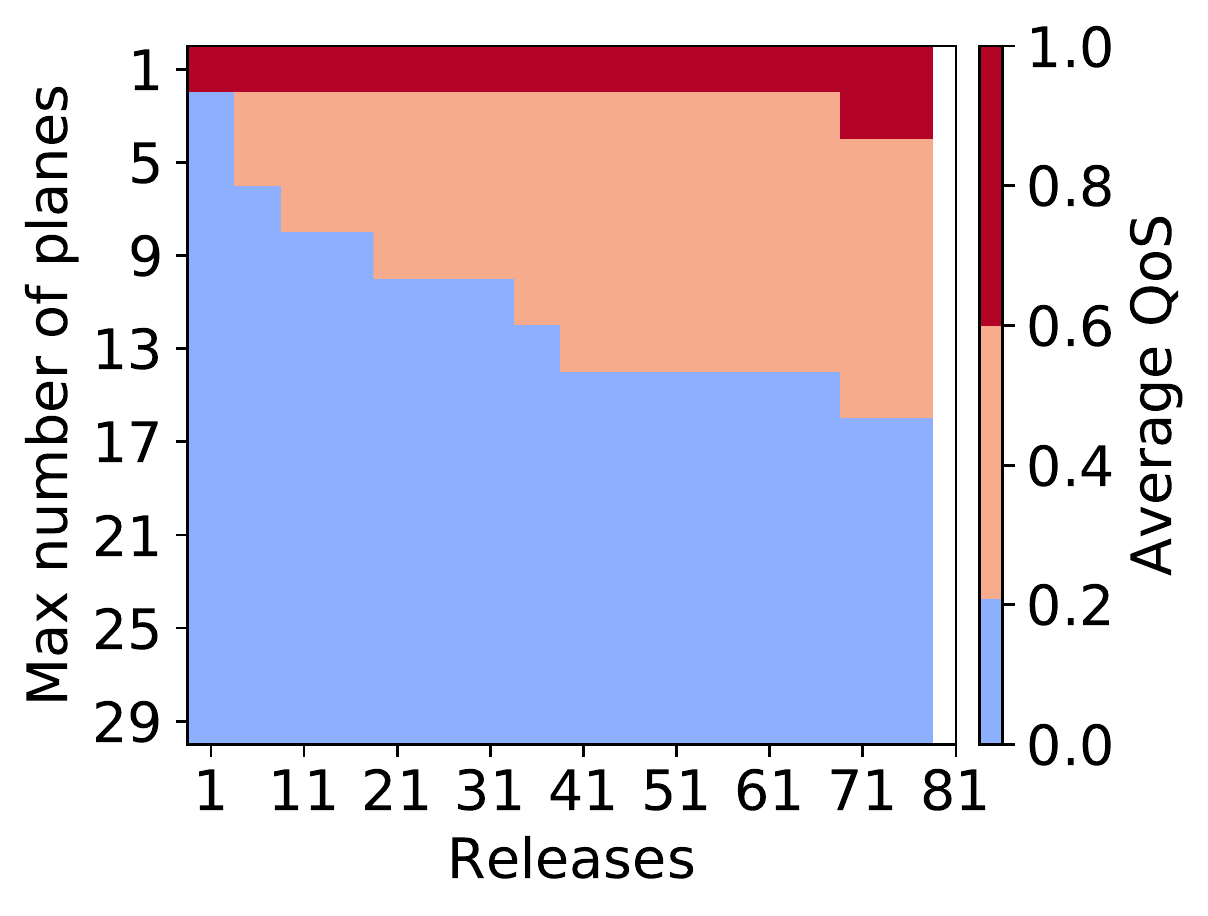}
		\caption{\small Heatmap ($r = 1.0$)}
			\label{fig:varying-planes-heatmap-utility-r1}
	\end{subfigure}
	\vspace{-2mm}
	\caption{\small Average QoS $Q$ of conservatively released planes over successive releasing}
	\label{fig:utility-conservative-releasing}
	\vspace{-2mm}
\end{figure}

Fig. \ref{fig:utility-conservative-releasing} shows the $Q$ values for conservatively released planes over successively released partial spaces. Here, regardless of $r$, we can see that $Q$ decreases as we increase $\mathtt{max\_planes}$. Consequently, if we fix $\mathtt{max\_planes}$, more successive partial releases increases $Q$. Intuitively, more information should, i.e. more releases, should decrease $Q$ but becasue we are also limiting the maximum allowable number of planes to be released, this introduces errors and thus increasing $Q$.

Let's look at the sample shown earlier in Fig. \ref{fig:protection-examples}. As shown in Fig. \ref{fig:partial-conservative-releases}, if we set $\mathtt{max\_planes} = 3$, this means that the succeeding releases will be forced on to these previously released 3 planes. Other structures present on these succeeding releases that cannot be subsumed by these planes will effectively not be released, and, thus, contribute to the QoS calculation. Example of these other structures are the planes shown in Fig. \ref{fig:partial-generalized-releases} but not present in Fig. \ref{fig:partial-conservative-releases}. Furthermore, the increasing $Q$ with increasing radius is also corroborated by Fig. \ref{fig:utility-conservative-releasing} which shows that the values for $r = 0.5$ are less than or equal to that of $r = 1.0$ for the same number of releases or $\mathtt{max\_planes}$.

\vspace{-5mm}\textit{\paragraph{\textbf{Takeaway}.} To achieve better QoS with generalized spaces, smaller size or radius is preferred. And, if conservative releasing is applied, good $Q$, i.e. $<0.2$, can be achieved with lower number of releases.}

\section{Discussion}\label{sec:discussion}

\paragraph*{\textbf{Challenge with point cloud data.}} Point clouds are unordered and variable-sized data structures which makes computation of, say, mutual information challenging. This poses a further challenge in directly applying well-studied privacy frameworks such as differentially private mechanisms. Thus, in this work, we measured privacy in terms of the errors of the two-level spatial inference attack described earlier.

\begin{figure}[t]
    \begin{subfigure}[]{0.47\columnwidth}
		\centering
			\includegraphics[width=\columnwidth]{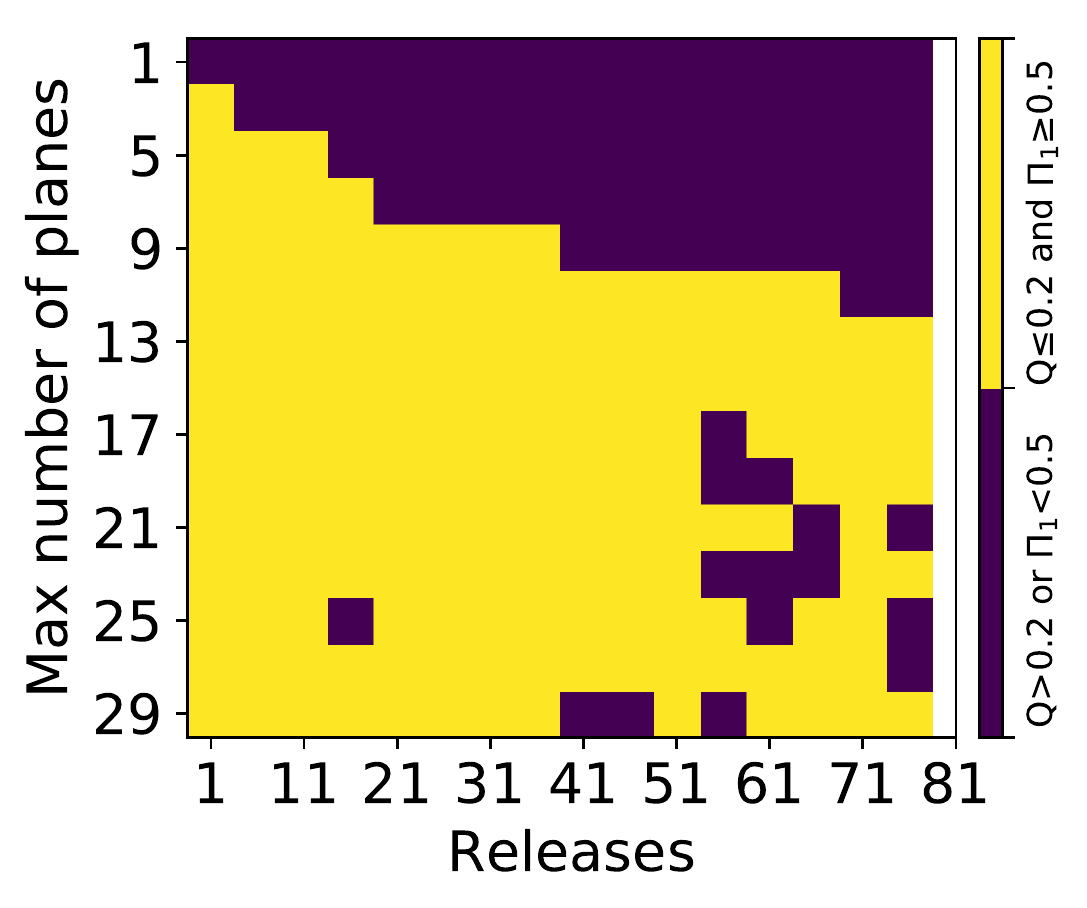}
		\caption{\small $r = 0.5$}
			\label{fig:varying-planes-comb-05}
	\end{subfigure}
    \begin{subfigure}[]{0.47\columnwidth}
		\centering
			\includegraphics[width=\columnwidth]{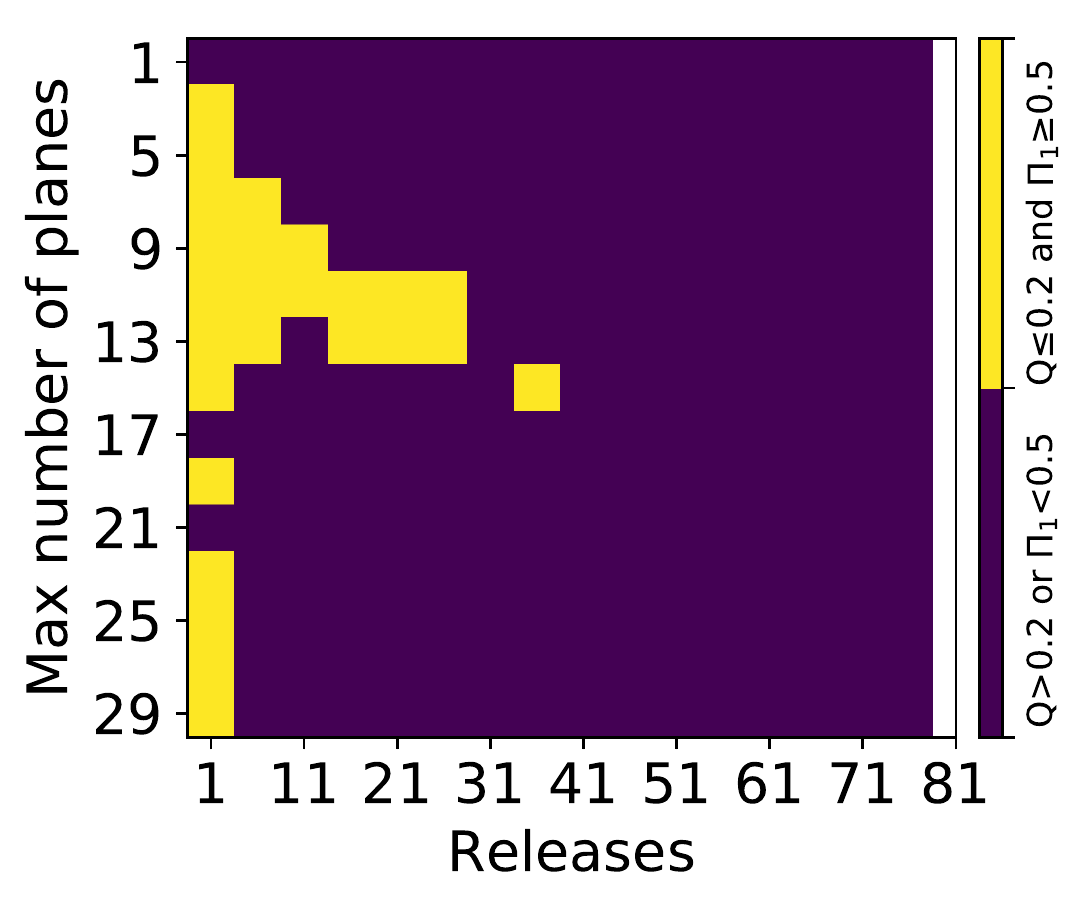}
		\caption{\small $r = 1.0$}
			\label{fig:varying-planes-comb-10}
	\end{subfigure}
	\vspace{-2mm}
	\caption{\small Intersection map of $Q\leq0.2$ and $\Pi_1\geq0.5$}
	\label{fig:utility-vs-privacy}
	\vspace{-2mm}
\end{figure}

\paragraph*{\textbf{Utility vs Privacy.}} Fig. \ref{fig:utility-vs-privacy} shows an intersection map for acceptable $Q$ and $\Pi_1$, i.e. $Q\leq0.2$ and $\Pi_1\geq0.5$. For $r = 0.5m$, as shown in Fig.\ref{fig:varying-planes-comb-05}, the good intersection regions are primarily dictated by good $Q$, i.e. up to 51 releases and up to 23 $\mathtt{max\_planes}$. Contrariwise, for $r=1.0m$, as shown in Fig.\ref{fig:varying-planes-comb-10}, the good intersection regions are very scarce and are primarily dictated by good $\Pi_1$; specifically, up to 13 $\mathtt{max\_planes}$ and up to 26 releases but lower $\mathtt{max\_planes}$ worsen $Q$. Thus, for $r = 1.0$, if we prioritize privacy over utility/QoS, then we can stick to $\mathtt{max\_planes} \leq 11$ but no limit in the number of successive releases. But, if we want better $Q$, we can reduce the size of the space, say, $r = 0.5m$, and enjoy more freedom in the number of releases.

\paragraph*{\textbf{Functional Utility.}} So far, we have only been focused on data utility. Specifically, we assume that as long as the revealed generalized space has a low $Q$, we hypothesize that the same original functionality can still be provided with the user perceiving minimal to no difference. For a great deal of MR applications requiring ``anchoring'' to surfaces, as long as the plane generalizations are kept aligned with their corresponding true surfaces, the user may indeed perceive minimal to no difference. For the case when a specific target, e.g. a 2D image marker or 3D object, is the desired anchor, the position (and orientation) of this target can be provided as either a small plane generalization or as a point anchor which can be used together with the other plane generalizations. 
However, it would also be worthwhile as future work to implement the proposed mechanism for protection over various MR applications and use cases, and measure actual user satisfaction or Quality-of-Experience (QoE).

\paragraph*{\textbf{Extensions to other 3D data sources.}} Furthermore, we have only focused on 3D point cloud data captured by MR platforms. However, there are other platforms and use cases in which 3D data is used. For example, 3D lidar data captured and used in geo-spatial work, by self-driving cars, and now by many other applications on recent smartphones. As we have mentioned, the methods we have used for spatial attacks are adapted from place recognition methods originally used for 3D lidar data. Now, conversely, the protection methods we have designed and developed can be applied on these other platforms. Of course, the figuring out a balance between functional utility, say, for a self-driving car, and privacy will still be a significant a challenge.


\paragraph*{\textbf{Improving the attackers.}} It is not the focus of this work to develop the best attacker. However, countermeasures are only as good as the best attack it can defend against. Thus, it is a worthwhile effort to explore improvements on the attack methods as future work such as an optimal submap generator for \texttt{pointnetvlad}. Likewise, the intrinsic descriptors used by \textsc{NN-matcher} can be transformed to a more discriminative global descriptor using \texttt{NetVLAD} to improve \textsc{NN-matcher}'s intra-space performance. Moreover, \texttt{pointnetvlad} can potentially be improved by replacing the \texttt{PointNet} layer with the updated \texttt{PointNet++} \cite{qi2017pointnet++}.

\section{Conclusion and Future Work}\label{sec:conclusion}
Currently, MR services are provided with full and indefinite access to the 3D spatial information, i.e. 3D point cloud data, captured by MR-capable devices. In this work, we highlight the risks associated to the indefinite access of applications to these 3D point cloud data. 
We have demonstrated how a descriptor-based and a deep neural network-based inferrers can be used for 3D spatial inference. Using these inferrers, we can recognize 
the current space of a user using the 3D point cloud information captured by their MR device. 
Furthermore, we have extended the capability of the inferrers to also reveal the exact location within the space 
of the user. Our validation results reveal that, unsurprisingly, raw 3D point cloud data readily reveals the specific location of spaces with average intra-space error of as low as $1 m$. And we reiterate that naive spatial generalizations, i.e. using the RANSAC plane-fitting algorithm, is still inadequate. 
Therefore, it is necessary to provide users with further protection. 

In this work, we have demonstrated that we can enhance privacy protection with plane generalizations by 
conservatively releasing the generalized planes provided to applications. Our experimental investigation over accumulated data from successive releases (emulating user movement) shows that we can reveal up to 11 planes and avoid inter-space inference at least for half of the time for large enough revealed spaces, i.e. $r\leq 1.0$. Specifically, with such very conservative releases, the success rate of any (inter-space) inferrer is no better than a random guess. And, for the occasions that the adversary correctly recognizes the space, the intra-space location can be off by at least 3 meters. Moreover, we quantified the privacy improvement in terms of both $\Pi_1$ and $\Pi_2$ by reducing the size of the partial spaces to be revealed. Consequently, in terms of data utility $Q$, a smaller size is preferred to provide good (i.e. low) $Q$. 

Overall, \textit{conservative releasing} is a viable solution for protecting spatial privacy of users as they use MR services while providing measurable privacy guarantees or thresholds. 
Plane generalization are already--or can easily be--implemented in most MR platforms. Thus, what remains is the implementation of conservative plane releasing to provide protection in these MR platforms. In our future work, we aim to develop a software library for mobile devices that includes the mechanisms proposed in this work which facilitates MR app designers to develop privacy-aware MR applications. 


\section*{Availability}
The source codes of our experiments is available at \url{https://github.com/spatial-privacy/3d-spatial-privacy}.

\section*{Appendix: The NN-matcher}

\begin{algorithm}[b]
    \scriptsize
	\KwData{\;
	$\{\{f_1,k_1\}, \{f_2,k_2\}, ... ,  \{f_I,k_I\}\}$ the ensemble set of information from reference spaces where $\{f_i,k_i\}$ is the set of key point-descriptor pairs for reference space $i$\;
	$\{f_q,k_q\}$ is the set of key point-descriptor pairs for a query space\;
	}
	\BlankLine

	\KwResult{$i_*$ candidate reference space, and $\{k_q \mapsto k_{1*}\}$ the matched query and reference key points}
	
	\BlankLine
	\DontPrintSemicolon

	\textbf{J} a 3D feature matcher, i.e. 2nn matcher\;
	\textbf{T} is NNDR threshold = 0.9\;
	\textbf{E}$_{i,q}$ a 2d-array of 2nn indices\; 
    \textbf{D}$_{i,q}$ a 2d-array of 2nn distances \tcp{both \textbf{E} and \textbf{D} have shape Q$\times$2, where Q is the length of $f_q$} 
	\textbf{S} = $\{s_1, s_2, ...\}$, the global matching scores of reference spaces $i$\;
	\textbf{K} = $\{\{k_q \mapsto k_1\},\{k_q \mapsto k_2\}, ... \}$ are the set-pairs of matched query key points $k_q$ with the key points $k_i$ of every reference space $i$

    \BlankLine

	\For{$\{f_i,k_i\}$ $\leftarrow$ $\{f_1,k_1\}$ \KwTo $\{f_I,k_I\}$}
	{
		$E_{i,q}, D_{i,q}$ = \textbf{J}($f_i, f_q$)\;
		$D_{i,q}$ $\leftarrow$ $\mathtt{max\_normalize}$($D_{i,q}$)\;
		$\mathbb{D} \leftarrow D_{i,q}[:,0]$/$D_{i,q}[:,1]$ \tcp{the array of the same length as $f_q$ containing the NNDR for every query feature with its top-2 matches among the reference features}
		(optional strict step; applied in \cite{deguzman2019firstlook})$E_{i,T} \subset E_{i,q}$, where $\mathbb{D}_1 <$ T \tcp{the subset of indices with NNDR<T}
		$\{k_{q} \subset k_q, k_{i} \subset k_i\}$, where $k_q \mapsto k_i$ \tcp{the unique key point matches}
		$E_{i} \subset E_{i,T}$ of $\{k_{q}, k_{i}\}$ \tcp{the subset indices corresponding to the unique key point matches}
		\textbf{K} $\leftarrow$ \textbf{K} + $\{k_{q}, k_{i}\}$ \;
		$s_i = (1-\mathtt{mean}(\mathbb{D}[E_{i}]))\cdot\frac{|E_{i}|}{|f_q|}$ \;
		\textbf{S} $\leftarrow$ \textbf{S} + $s_i$\;
	}

	$i_*=$ \texttt{argmax}(\textbf{S}) \;
	\caption{Inter-space matching algorithm}
	\label{alg:NN-matcher}
\end{algorithm}

\subsubsection*{Intrinsic 3D descriptors} Features that describe and discriminate among 3D spaces are usually used for inference modelling, and there are considerable features in 3D point clouds for it to be directly used as a 3D descriptor. However, point cloud extractions of the same space may produce translated and/or rotated versions of each other; thus, for a machine to recognize that these point clouds are of the same space, machine understanding has to be translation- and rotation-invariant. To provide invariance, we utilize an existing 3D description algorithm called \textit{spin image} (SI) descriptors \cite{johnson1999using}. The SI description algorithm only uses the normal vectors associated with the points. Moreover, it is not reliant on curvature or any other second-order or higher-order structural information. 

The original spin image descriptor algorithm does not have a key point selection process. For our implementation, we implement a pseudo key point selection process by using a sub-sampled point cloud space (sampling factor of 5) as our chosen key points instead of the complete point cloud. This reduces the likelihood of having neighbouring points having exactly the same descriptors. Then, we compute the spin image descriptors made up of cylindrical bins relative to a local cylindrical coordinate system around each key point. The \textit{spin} comes from the fact that spinning the key point about its normal will have no effect on the computed descriptor; thus, it is rotation-invariant about its normal. Consequently, the spinning effect reduces the impact of variations within that spin which makes SI descriptors more robust compared to other descriptors which are reliant on second-order information such as curvatures. Furthermore, as described in \S\ref{subsec:generalization}, plane generalization removes curvatures as well as other second- and higher-order information which makes its use as a geometric description information impractical. It has been shown that, indeed, SI descriptors are more robust when spaces are plane generalized than a similar algorithm, i.e. self similarity \cite{huang2012point}, which requires curvature information \cite{deguzman2019firstlook}. 

\vspace{1mm}
\noindent{\emph{Inter-space inference.}} 
A matching function maps two sets of features $\{f_{a,0},f_{a,1},...\}$ and $\{f_{b,0},f_{b,1},...\}$ of spaces $S_a$ and $S_b$, respectively. To accept a match, we get the \textit{nearest neighbor distance ratio} (NNDR) of the features like so: 
$\frac{d(f_{a,1}, f_{b,1})}{d(f_{a,1}, f_{b,2})} < \text{threshold},$
where $d(\cdot)$ is some distance measure in the feature space, and descriptor $f_{b,1}$ is the nearest neighbor of descriptor $f_{a,1}$ while $f_{b,2}$ is the second nearest neighbor. If the NNDR falls below a set threshold, then $\{k_{a,1},f_{a,1}\}$ and $\{k_{b,1},f_{b,1}\}$ is an acceptable key point-feature pairs.

For our implementation, we use the Euclidean nearest neighbors, i.e. 2-$\mathtt{nn}$, for the feature distance function $d(\cdot)$. Afterwards, the NNDR of the two best matching reference descriptors for every query descriptor from the reference spaces are computed. Then, to determine the best matching space for inter-space inference, 
a weighted score is computed based on the product on the product of the percentage of uniquely indexed lowest ratio matches per candidate space, and that of the difference from 1 of the mean of the NNDRs of the unique (key point-wise) matches. The reference space, with the highest score--akin to the $\mathtt{argmax}$ operation of Eq. \ref{eq:inference}a--is the best candidate space for the query space. In this work, we utilized a relaxed version of the inference method utilized in
\cite{deguzman2019firstlook} which originally only accepts the uniquely matched key point pairs with NNDR $< 0.9$ before computing the weighted score. Alg. \ref{alg:NN-matcher} presents the pseudo-code for inter-space inference.

To determine the best matching space for inter-space inference, we, 
first, get the 2 nearest neighbors of a query feature among the reference features. Line 8 in both Alg. \ref{alg:NN-matcher} shows this $2nn$ distance computation in the feature space which produces two arrays: \textbf{E}$_{i,q}$ contains indices of the $2nn$ descriptors from reference space $i$, while \textbf{D}$_{i,q}$ contains the distance values. Afterwards, \textbf{D}$_{i,q}$ is maximum-normalized (line 9) before we compute the NNDR (line 10). Originally, in \cite{deguzman2019firstlook}, an optional step (line 11) trims the $2nn$ matches and only keeps those below the set threshold \textbf{T}. Then, 
the set of matches are further trimmed (line 12, and 13) by only accepting the matches with unique key point matches; that is, query key points are matched to unique reference key points. If there are query key points matched to the same reference key points, we keep the pair with the lowest NNDR. The resulting matches are kept (line 14), while the global matching score is computed as shown in line 15. The reference space with the highest score is picked as the inter-space candidate (line 17). The key point pairs of the resulting matched space is provided to the intra-space matcher to determine the intra-space location of the user.

\begin{algorithm}[b]
	\scriptsize
	\DontPrintSemicolon
	
	\KwData{$\{k_q \mapsto k_{*}\}$ the matched query and reference key points\;
		$\mathbb{D}_*$ the NNDR ratios of the  matched key points
	}
	\BlankLine
	
	\KwResult{$c_*$ centroid of the matched reference key points $k_{i*}$}
	
	\BlankLine
	
	$G(\cdot)$ a graph\;
	$T_1$ is NNDR threshold = 0.9\;
	$T_2$ is geometric similarity threshold = 0.95\;
	
	\BlankLine
	
	$\{\mathtt{k_{q,1}} \subset k_q, \mathtt{k_{i*,1}} \subset k_{i*}\}$, where $\mathbb{D}_* <T_1$ \tcp{For the Strict-NN, this step has already been performed}
	$G_{q,1}$ = $G(\mathtt{k_{q,1}})$ \tcp{complete graph with elements of key point positions $p_k \in \mathtt{k_{q,1}} (\in \text{\textbf{R}}^3)$ as vertices, and edges defined by the point-to-point distance vector $\hat{v} \in \text{\textbf{R}}^3$.}
	$G_{*,1}$ = $G(\mathtt{k_{*,1}})$ \tcp{we get the same for the reference key points; in matrix form, the dimensions of $G_{q,1}$ and $G_{*,1}$ are the same and their elements should be consistently ordered following their matched key points}
	$V_q \leftarrow ||\text{Edges}(G_{q,1})||_{L2},\newline V_* \leftarrow ||\text{Edges}(G_{*,1})||_{L2}$ \tcp{we get the L2-norms of the distance vectors which are the edges of the complete graphs}
	$A_q \leftarrow \mathtt{internal\_angles}(\text{Edges}(G_{q,1})),\newline A_* \leftarrow  \mathtt{internal\_angles}(\text{Edges}(G_{*,1}))$ \tcp{we get the internal angles formed by the distance vectors with each other}
	$S_d = \mathtt{exp}(-0.5*|V_q - V_*|)$ \tcp{we compute the distance similarity using an exponential function with rate = $-0.5$; $S_d \in [0,1]$ with 1 being the highest similarity}
	$S_{\phi} = \mathtt{cosine\_similarity}(A_q, A_*)$ \tcp{we compute the cosine similarity of the two sets of internal angles; $S_\phi \in [0,1]$ with 1 being the highest similarity}
	$S = S_d\cdot S_\phi$ \tcp{combined product similarity measure}
	$\{\mathtt{k_{q,2}} \subset \mathtt{k_{q,1}}, \mathtt{k_{*,2}} \subset \mathtt{k_{*,1}}\}$, where $S\geq T_2$ \tcp{only keep the key point pairs with $S\geq T_2$}
	$c_*$ = $\mathtt{centroid}$($\mathtt{k_{*,2}}$)
	\caption{Intra-space matching}
	\label{alg:intra-space-matching}
\end{algorithm}

\subsubsection*{Intra-space inference}

The resulting keypoint matches are fed to the intra-space matcher as shown in Alg. \ref{alg:intra-space-matching} to determine the intra-space location of the user. We trim the key point pairs by only accepting those whose feature NNDR are below the set threshold $T_1$ (line 4). Then, we find the fully-connected graphs of both query (line 5) and matched reference (line 6) key points; the key points are the vertices while the edges are defined by the point-to-point vector connecting the key points. We compute the distances using L2-norm of the point-to-point distance vectors (line 7). We also compute the internal angles (using cosine similarity) formed by the vectors (line 8). We compute a distance similarity measure (line 9) $S_d$, and an angular similarity measure (line 10) $S_\phi$. The product of the two measures forms the combined similarity metric $S$ (line 11). The vertices, i.e. key point pairs, with an acceptable $S$, i.e. $\geq 0.95$, are the accepted key point pairs.The intra-space location is the centroid of the accepted matched reference key points.

\bibliographystyle{ACM-Reference-Format}
\bibliography{bib_all.bib}

\end{document}